\documentclass[10pt,twocolumn,letterpaper]{article}

\usepackage{cvpr}              % To produce the CAMERA-READY version
\usepackage{graphicx}
\usepackage{amsmath}
\usepackage{amssymb}
\usepackage{enumitem}
\usepackage{subcaption}	% You need this package for subfigures
\usepackage{booktabs}
\usepackage{multirow}
\usepackage{xspace}
\usepackage{color, colortbl}
\definecolor{Gray}{gray}{0.9}

\def\*#1{\mathbf{#1}}
\usepackage{algorithm}
\usepackage[ruled,vlined,algo2e]{algorithm2e}
\newcommand{\model}{\texttt{STUD}\xspace}

\usepackage[pagebackref,breaklinks,colorlinks]{hyperref}

\usepackage[capitalize]{cleveref}
\crefname{section}{Sec.}{Secs.}
\Crefname{section}{Section}{Sections}
\Crefname{table}{Table}{Tables}
\crefname{table}{Tab.}{Tabs.}

\def\cvprPaperID{5517} % *** Enter the CVPR Paper ID here
\def\confName{CVPR}
\def\confYear{2022}

\begin{document}


\title{Unknown-Aware Object Detection:\\ Learning What You Don't Know from Videos in the Wild}

\author{Xuefeng Du$^{1}$, Xin Wang$^{2}$, Gabriel Gozum$^{1}$, and Yixuan Li$^{1}$\\
% Department of Computer Sciences \\
$^{1}$University of Wisconsin-Madison, $^{2}$Microsoft Research \\
{\tt\small \{xfdu,sharonli\}@cs.wisc.edu, wanxin@microsoft.com, ggozum@wisc.edu}
}
\maketitle

%%%%%%%%% ABSTRACT
\begin{abstract}
Building reliable object detectors that can detect out-of-distribution (OOD) objects is critical yet underexplored. One of the key challenges is that models lack supervision signals from unknown data, producing overconfident predictions on OOD objects. 
% Collecting and annotating OOD objects in complex visual scenes can be costly and sometimes infeasible in practice. 
%Previous approaches rely on real outlier datasets for regularization, which can be costly and sometimes  infeasible  to  obtain  in  practice.  
% This limitation can be even severe for more complicated object detectors since detecting unknowns for object detection requires a finer-grained understanding of complex scenes and images. 
We propose a new unknown-aware object detection framework through Spatial-Temporal Unknown Distillation (\model), which distills unknown objects from videos in the wild and meaningfully regularizes the model’s decision boundary. 
%\model exploits the rich information in videos, which naturally  encapsulate  a  mixture  of  both  known  and  unknown objects.\@
% leverages the {\rm temporal correspondence} property of the sequential videos to
% distills the unknown objects from known videos which can meaningfully regularize the model’s decision boundary during training.
% In  this  work,  we  \modely the effectiveness of the sequential video data to mitigate this issue. We find that
% the very act of {\rm looking for correspondences} in sequential videos enables us to identify the unknown objects that can meaningfully regularize the model’s decision boundary during training. 
%\widehat
\model first identifies the unknown candidate object proposals in the spatial dimension, and then aggregates the candidates across multiple video frames to form a diverse set of unknown objects near the decision boundary. 
% and synthesizes a diverse set of unknown objects by combining object features in several reference frames of a video.
% matching each object to the distant object proposals across multiple frames in a video and then synthesizing the unknown object features by weighting these proposals with the matching score. 
Alongside, we employ an energy-based uncertainty regularization loss, which contrastively shapes the uncertainty space between the in-distribution and distilled unknown objects. \model establishes the state-of-the-art performance on OOD detection tasks for object detection, reducing the FPR95 score by over 10\% compared to the previous best method. Code is available at \url{https://github.com/deeplearning-wisc/stud}.
\end{abstract}

%%%%%%%%% BODY TEXT
\section{Introduction}

\label{sec:intro}
Object detection models have achieved remarkable success in known contexts for which they are trained. Yet, they often struggle with out-of-distribution (OOD) data---samples from unknown classes that the network has not been exposed to during training, and therefore should not be predicted by the model in testing.  Teaching the object detectors to be aware of unknown objects is critical for building a reliable vision system,  especially in safety-critical applications like autonomous driving~\cite{du2022vos} and medical analysis~\cite{DBLP:journals/corr/abs-2007-04250}. 

% In particular, modern neural networks deployed in real-world systems can encounter out-of-distribution (OOD) inputs---unknown samples that the network has not been exposed to during training, and therefore should not be predicted by the model in testing. 
%In particular, neural networks can produce overconfident predictions for OOD data~\cite{nguyen2015deep}.
%Object detection models can undesirably inherit these issues. 

\begin{figure}
    \centering
    \vspace{-1em}
    \includegraphics[width=1.0\linewidth]{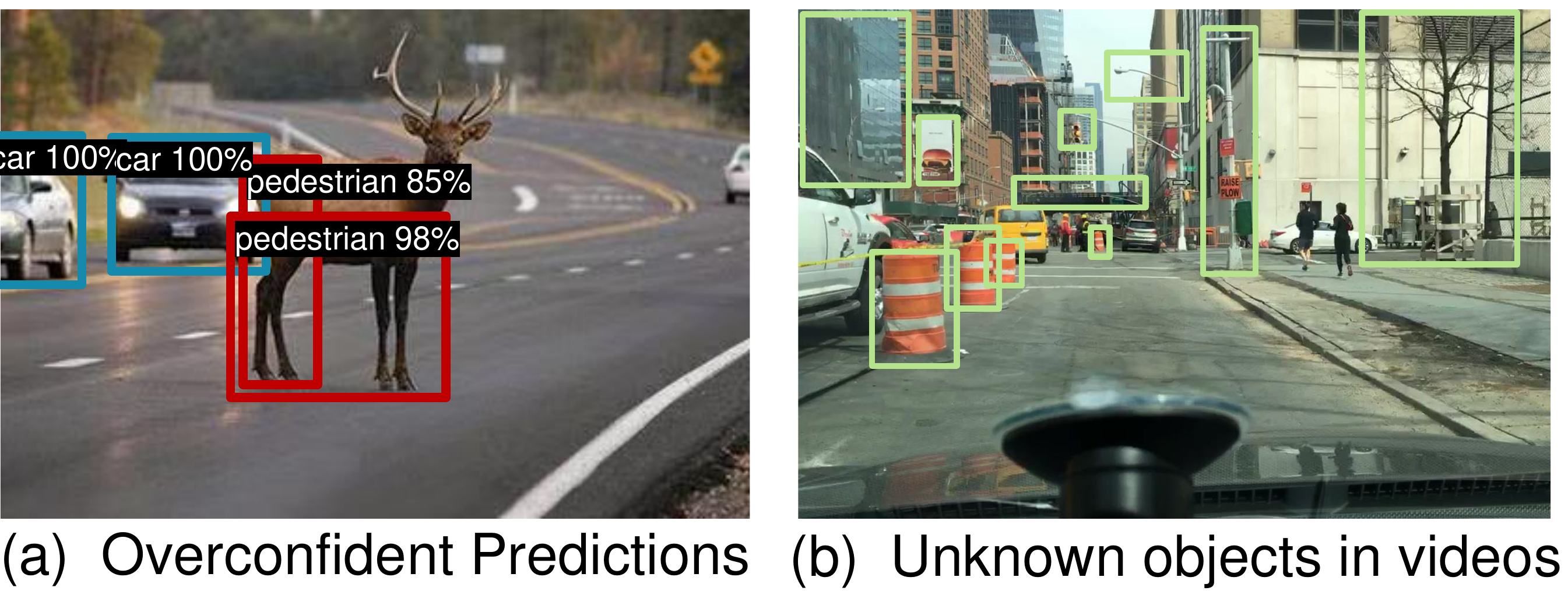}
    \caption{\small (a) Vanilla object detectors can predict OOD objects (\emph{e.g.}, deer) as an ID class (\emph{e.g.}, pedestrian) with high confidence. (b) Unknown objects (in bounding boxes) naturally exist in the video datasets, such as billboards, traffic cones, overbridges, street lights, etc. Image is taken from the BDD100K dataset~\cite{DBLP:conf/cvpr/YuCWXCLMD20}.  }
    \label{fig:teaser}
    \vspace{-1.5em}
\end{figure}

While much research progress is made in OOD detection for classification models~\cite{hendrycks2016baseline,lakshminarayanan2017simple, liang2018enhancing, lee2018simple,liu2020energy,tack2020csi, hsu2020generalized}, the problem remains underexplored  in the context of object detection. Unlike image-level OOD detection, detecting unknowns for object detection requires a finer-grained understanding of the complex scenes. In practice, an image can be OOD in specific regions while being in-distribution (ID) elsewhere. Taking  autonomous driving as an example, we observe that an object detection model trained to recognize ID objects (\emph{e.g.,} {cars}, {pedestrians}) can produce a high-confidence prediction for an unseen object such as a {deer}; see Figure~\ref{fig:teaser}(a). This happens when our object detector  minimizes its training error {without explicitly accounting for the uncertainty that could appear outside the training categories}. Unfortunately, the plethora of ways that unknown objects can emerge are innumerable in an open world. It is arguably expensive to annotate a large number of OOD objects in complex scenes---in addition to the already costly process of ID data collection. %Moreover, one cannot anticipate all possible unknown objects in advance, especially in high-dimensional visual space. 
%While one could annotate OOD objects, such a solution can be arguably expensive.%  and meanwhile lacks flexibility, as we need to re-annotate the dataset if the ID data changes or expands. 

%Such a failure case gives rise to the importance of estimating OOD uncertainty for safety-critical scenarios. 

In this paper,\@
% we propose a novel \emph{unknown-aware object detection} framework through spatial-temporal unknown distillation (\model), which automatically generates unknown objects guided by the rich and diverse information in videos.  
we propose a new \emph{unknown-aware object detection} framework through Spatial-Temporal Unknown Distillation (\model), which distills unknown objects from videos in the wild and meaningfully regularizes the model’s decision boundary.\@
Video data naturally captures the open-world environment that the model operates in, and encapsulates a mixture of both known and unknown objects; see Figure~\ref{fig:teaser}(b). For example, {buildings} and {trees} (OOD) may appear in the driving video, though they are not labeled explicitly for training an object detector for cars and pedestrians (ID).  Our approach draws an analogy to the concept of distillation in chemistry, which refers to the ``{process of separating the substances from a mixture}''~\cite{doi:10.1021/ie50303a003}. While classic object detection models primarily use the labeled known objects for training, we attempt to capitalize on the unknown ones for model regularization by jointly optimizing object detection and OOD detection performance. %While the intuition is straightforward, the key question is how to automatically identify and construct these unknown objects at the proper level of difficulty and diversity, without explicit supervision signal of unknown objects. %Simple solution such as background proposals~\cite{DBLP:journals/corr/abs-2103-02603} are too trivial to generalize to realistic OOD objects.  

 Concretely, our framework consists of two components, tackling challenges of (\textbf{1}) distilling diverse unknown objects from videos, and (\textbf{2}) regularizing object detector with the distilled unknown objects. To address the first problem, we introduce a new \emph{spatial-temporal unknown distillation} approach, which automatically constructs \emph{diverse}  unknown objects (Section~\ref{sec:unknown_distill}). In the spatial dimension, for each ID object in a frame, we identify the unknown object candidates in the reference frames based on an OOD measurement. We then distill the unknown object by linearly combining the selected objects in the feature space, weighted by the dissimilarity measurement. The distilled unknown object therefore captures a more {diverse} distribution over multiple objects than using single ones. In the temporal dimension, we propose aggregating unknown objects from multiple video frames, which captures additional diversity of unknowns in the temporal dimension.   %\texttt{\model} trades off exploiting the video consistency and exploring the unknown diversity by using nearby reference frames, where the effect of the adjacency level is ablated in Section~\ref{sec:ablation}.

Leveraging the distilled unknown objects, we further employ an unknown-aware training objective (Section~\ref{sec:regularization}). Unlike vanilla object detection, we train the object detector with an uncertainty regularization branch. Our regularization facilitates learning a more conservative decision boundary between ID and OOD objects, which helps flag unseen OOD objects during inference. To achieve this, the regularization contrastively shapes the uncertainty surface, which produces larger probabilistic scores for ID objects and vice versa, enabling effective OOD detection in testing. %, amplifying the separability between known ID objects and distilled unknowns.  
%During training, \texttt{\model} simultaneously performs the object detection task as well as the OOD uncertainty regularization. 
%The uncertainty estimation branch 
% Furthermore, we show that xxx benefits from the temporal relationship aggregation between proposals from multiple frames, which potentially enables a diverse unknown distillation procedure than relying on the spatial correlation between two frames (Section xxx). 
Our key contributions are summarized as follows:
  \vspace{-0.5em}
\begin{itemize}[leftmargin=1em]
% \vspace{-0.2em}
    \item We propose a new framework \texttt{\model}, addressing a challenging yet underexplored  problem of unknown-aware  object detection. To the best of our knowledge, we are the first to exploit the rich information from videos to enable OOD identification for the object detection models.% to favorably identify the OOD objects during inference.
    % Our effort facilitates future research to evaluate OOD detection in a setting closer to real-world applications.
      \vspace{-0.5em}
%   \item  We conduct extensive ablations and reveal important insights by contrasting different unknown object distillation approaches. We show that \texttt{\model} is more advantageous than synthesizing unknowns directly in the high-dimensional pixel space (e.g., using GAN~\cite{lee2018training}) or using negative proposals as unknowns~\cite{DBLP:journals/corr/abs-2103-02603}.
   \item  \texttt{\model} effectively
   regularizes object detectors by distilling diverse unknown objects in both spatial and temporal dimensions without costly human annotations of OOD objects. Moreover, we show that \texttt{\model} is more advantageous than synthesizing unknowns in the high-dimensional pixel space (e.g., using GAN~\cite{lee2018training}) or using negative proposals as unknowns~\cite{DBLP:journals/corr/abs-2103-02603}.
  \vspace{-0.5em}
   \item We extensively evaluate
  the proposed \texttt{\model} on large-scale BDD100K~\cite{DBLP:conf/cvpr/YuCWXCLMD20} and Youtube-VIS datasets~\cite{DBLP:conf/iccv/YangFX19}.  \texttt{\model}  obtains state-of-the-art results, outperforming the best baseline by a large margin (10.88\% in FPR95 on BDD100K) while preserving the accuracy of object detection on ID data. 
\end{itemize}
\vspace{-1em}

\section{Problem Setup}

\begin{figure*}[t]
    \centering
    \includegraphics[width=1.0\linewidth]{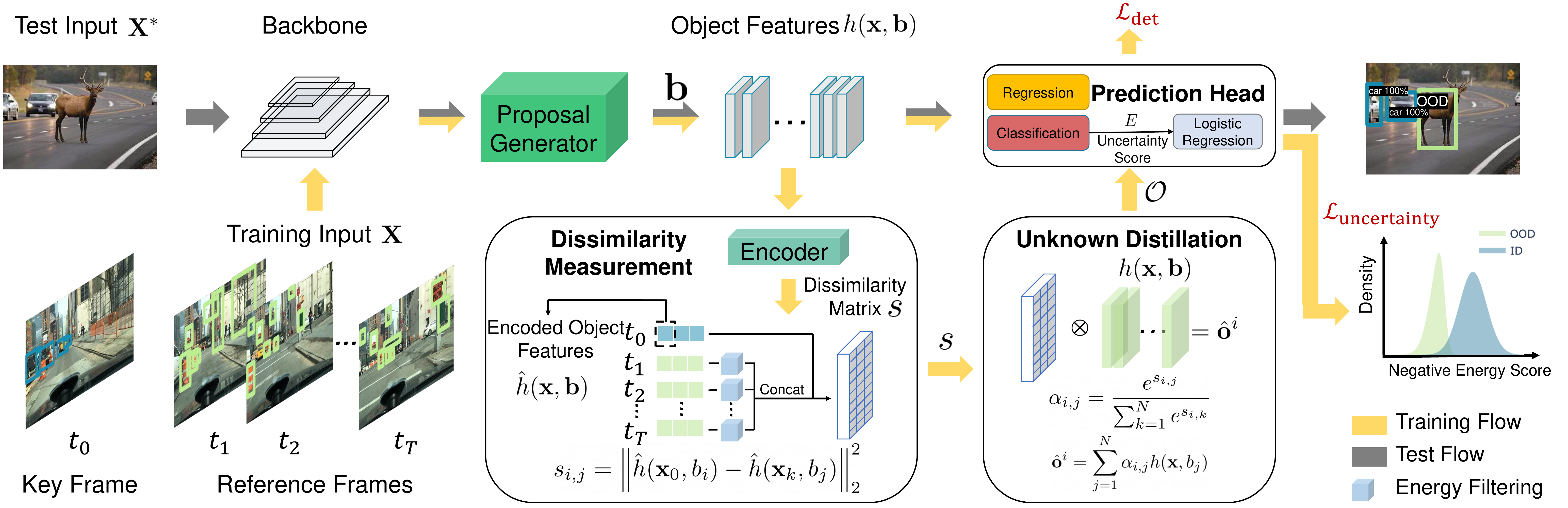}
    \vspace{-2em}
    \caption{\small \textbf{Overview of the  proposed unknown-aware object detection framework $\model$.} For an ID object from the key frame encoded as $\hat{h}(\mathbf{x}_0,\mathbf{b}_i)$, we perform energy filtering to identify the unknown object candidates in the reference frames. We then distill the unknown object  $\hat{\mathbf{o}}^{i}$ by linearly combining the unknown objects in the feature space, weighted by the dissimilarity score $s_{i,j}$. The distilled unknowns, along with the ID objects, are used to train the uncertainty regularization branch ($\mathcal{L}_{\text{uncertainty}}$). $\mathcal{L}_{\text{uncertainty}}$ contrastively shapes the uncertainty surface, which produces a larger score for ID objects and vice versa. During testing, we use the output of the logistic regression for OOD detection. $\otimes$ denotes the operation in Equation~\eqref{eq:synthesis} and $1\leq k \leq T$ is the index of the reference frames. }
    \label{fig:framework}
    \vspace{-1.2em}
\end{figure*}
We start by formulating the OOD detection problem for the object detection task. Most previous formulations of OOD detection treat entire images as anomalies, which can lead to ambiguity shown in Figure~\ref{fig:teaser}(a).  In particular, natural images are not monolithic entities but instead are composed of numerous objects and components.  Knowing which regions of an image are anomalous allows for the safe handling of unfamiliar objects.  Compared to image-level OOD detection, object-level OOD detection is more relevant in realistic perception systems, yet also more challenging as it requires reasoning OOD uncertainty at the fine-grained object level. We design reliable object detectors that are aware of unknown OOD objects in testing. That is, an object detector trained on the ID categories (\emph{e.g.}, cars, trucks) can identify test-time objects (\emph{e.g.}, deer) that do not belong to the training categories and refrain from making a confident prediction on them.
\vspace{-1em}
\paragraph{Setup.} We denote the input and label space by $\mathcal{X}=\mathbb{R}^d$ and $\mathcal{Y}=\{1,2,...,K\}$, respectively. Let $\mathbf{x} \in \mathcal{X}$ be the input image, $\mathbf{b} \in \mathbb{R}^4$ be the bounding box coordinates associated with objects in the image, and $y \in \mathcal{Y} $ be the semantic label of the object.  An object detection model is trained on ID data $\mathcal{D}=\left\{\left(\mathbf{x}_{i},\mathbf{b}_i, {y}_{i}\right)\right\}_{i=1}^{M}$
drawn from an unknown joint distribution $\mathcal{P}$. We use neural networks with parameters $\theta$ to model the bounding box regression $p_\theta(\mathbf{b} |\mathbf{x})$ and the classification $p_\theta(y|\mathbf{x},\mathbf{b})$.
%$\mathbf{b}_i$ denotes a set of $K$ ground truth object instances $\mathbf{b}_i=\{b_1,...,b_K\}$ and $\mathbf{y}_i=\{1,2,...,C\}$ denotes their class labels, where $y\in \left[1,2,...,\mathrm{C}\right]$. $b=\left[X, Y, W, H\right]$ which contains the center coordinates, the width and height of the bounding box, respectively. The goal of object detection is to predict the class labels and locations of the objects as well as possible.

\vspace{-1em}
\paragraph{OOD detection for object detection.} The OOD detection can be formulated as a binary classification problem, distinguishing between the in vs. out-of-distribution objects. Let $P_{\mathcal{X}}$ denote the marginal probability distribution on $\mathcal{X}$. 
Given a test input $\mathbf{x}^*\sim P_\mathcal{X}$, as well as an object $\mathbf{b}^*$ predicted by the object detector, the goal is to predict $p_\theta(g \vert \mathbf{x}^*, \mathbf{b}^*)$. We use $g=1$ to indicate a detected object being ID, and $g=0$ being OOD, with semantics outside the support of $\mathcal{Y}$.

\section{Unknown-Aware Object Detection}
% \paragraph{Overview.}
Our unknown-aware object detection framework trains an object detector in tandem with the OOD uncertainty regularization branch. Both share the feature
extractor and the prediction head and are jointly trained from scratch (see Figure~\ref{fig:framework}). Our framework encompasses two novel components, which address: (1) {how to distill diverse  unknown objects in the spatial and temporal dimensions} (Section~\ref{sec:unknown_distill}), and (2) {how to leverage the unknown objects for effective model regularization} (Section~\ref{sec:regularization}).
% (3) how to aggregate the temporal information for improved OOD detection (Section~\ref{sec:multi}) 
%and (3) how to perform OOD detection during inference time (Section~\ref{sec:inference})?

% \paragraph{Base Object detector.}
% We adopt the two-stage object detector, Faster R-CNN~\cite{ren2015faster}, as the base object detector, which
% can be replaced with other off-the-shelf detectors. A typical
% region-based object detector is composed of three components. A backbone feature extractor $\mathcal{F}$ is used to obtain
% image features. A set of region proposals is obtained from
% a region proposal network (RPN). Through ROI Align~\cite{ren2015faster},
% we obtain a set of ROI features, which are then fed into the
% box predictors for box classification and localization.

% xxx uses the region features produced by the RPN
% network as input for unknown distillation and soft outlier synthesis while other synthesis approaches, such as training generative models, i.e. GANs~\cite{goodfellow2014generative,lee2018training}, synthesizes images in the high-dimensional pixel space, which can be difficult to optimize.  The objective function to train
% the entire model is defined as
% \begin{equation}
%     \min \mathcal{L} = \min [\mathcal{L}_{det} +\beta \cdot \mathcal{L}_{uncertainty}].
% \end{equation}
% where $\beta$ is the scaling factor when combining the detection
% loss $\mathcal{L}_{det}$ and the uncertainty regularization loss $\mathcal{L}_{uncertainty}$.

\subsection{Spatial-Temporal Unknown Distillation}
\label{sec:unknown_distill}

Our approach \texttt{\model} distills unknown objects guided by the rich spatial-temporal information in videos, without explicit supervision signals of unknown objects.
% Our approach can draw analogy to the concept of {distillation} in chemistry, which refers to the ``\emph{process of separating the components or substances from a mixture}''~\cite{doi:10.1021/ie50303a003}.
Video data naturally encapsulates a mixture of both known and unknown objects. While classic object detection models primarily use the labeled known objects for training, we attempt to capitalize on the unknown ones for model regularization. For this reason, we term our approach \emph{unknown distillation}---extracting unknown objects \emph{w.r.t} the known objects. Notably, our distillation process for object detection is performed at the object level, in contrast to constructing the image-level outliers~\cite{hendrycks2018deep}. That is, for every ID object in a given frame, we construct a corresponding OOD counterpart. The distilled unknowns will be used for model regularization (Section~\ref{sec:regularization}).

\begin{figure}
    \centering
    \includegraphics[width=1.0\linewidth]{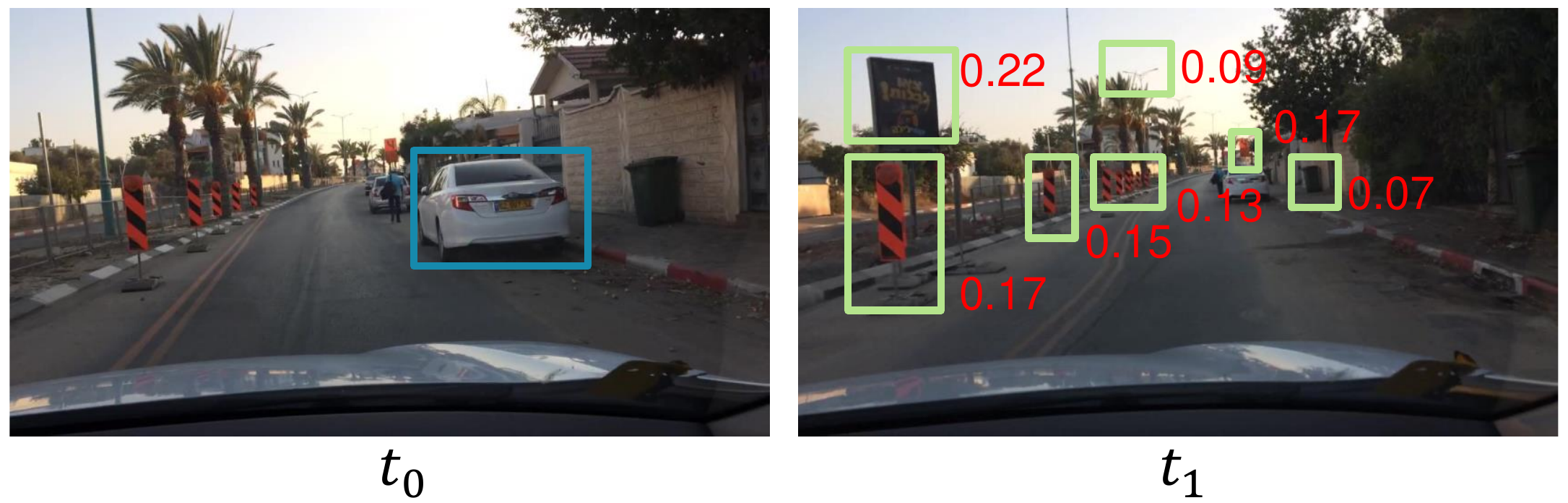}
    \vspace{-2em}
    \caption{\small  \textbf{The dissimilarity measurement.} For each ID object at timestamp $t_0$ (in blue), we discover the objects in the reference frame that are dissimilar to it (in green), which are more likely to contain OOD objects for model regularization. The red numbers show the dissimilarity after normalization (Equation~\eqref{eq:normalization}). 
    }
    \label{fig:matching}
    \vspace{-1.2em}
\end{figure}

While intuition is straightforward, challenges arise in constructing unknown objects in an unsupervised manner. The plethora of ways that unknown objects can emerge are innumerable in high-dimensional space. Taking the ID object \texttt{car} as an example (\emph{c.f.}~Figure~\ref{fig:matching}), the objects such as {billboards}, {trees}, {buildings}, etc. can all be considered as unknowns \emph{w.r.t} the car. This undesirably increases the sample complexity and demands a diverse collection of unknown objects to be observed. We tackle the challenge through distilling diverse unknown objects by leveraging the rich information in the spatial and temporal dimensions of videos. 
%While a straightforward idea is to train generative models such as GANs~\cite{goodfellow2014generative, lee2018training}, synthesizing images in the high-dimensional {pixel space} can be intractable to optimize. Our key idea is to synthesize unknown instances in the \emph{feature space}, which is more tractable given lower dimensionality. Moreover, our method is based on a discriminatively trained classifier  in  the  object  detector,  which  circumvents  the  difficult  optimization  process  in  training generative models.
% In the spatial dimension, given a ID object in a frame, \model synthesizes unknown instances by linearly combining  instance features  in  the  reference  frames using the calculated dissimilarity metric, which favorably increases the diversity of the synthesized unknowns compared to using independent reference proposals.

\vspace{-0.4cm}
\paragraph{Spatial unknown distillation.}  In the \emph{spatial} dimension, for each ID object in a given frame, we create the unknown counterpart through a linear combination of the object features from the reference frames, weighted by the dissimilarity measurement. Utilizing multiple objects captures a more diverse distribution of unknowns than using single ones. \texttt{\model} operates on the feature outputs from the proposal generator to calculate dissimilarity. Specifically, we consider 
a pair of frames $\mathbf{x}_0, \mathbf{x}_1$ at timestamps $t_0$ and $t_1$, designated {key frame} and {reference frame}, respectively. For an object $(\mathbf{x}, \mathbf{b})$,  we denote its feature representation as $h(\mathbf{x}, \mathbf{b}) \in \mathbb{R}^m$, where $m$ is the feature dimension. We collect a set of
object features $\{h(\mathbf{x_0}, b_i)\}_{i=1}^{N_0}$ and $\{h(\mathbf{x_1}, b_j)\}_{j=1}^{N_1}$ with the objectiveness score above a
threshold. We adopt a dissimilarity measurement using the
$L_2$ distance between two features:
\vspace{-0.5em}
\begin{equation}
\label{eq:distance}
    s_{i, j}=\left\|\hat{h}(\mathbf{x}_0, \mathbf{b}_i)-\hat{h}(\mathbf{x}_1, \*b_j)\right\|_{2}^{2},
    \vspace{-0.4em}
\end{equation}
where $\hat{h}(\mathbf{x_0}, \*b_i)$ and $\hat{h}(\mathbf{x_1}, \*b_j)$ are encoded
feature vectors obtained by a small network using the object features $h(\mathbf{x},\*b)$ as input. In our experiments, the
encoder consists of two convolutional layers with kernel size
of 3 × 3 and an average pooling layer. The larger $s_{i,j}$ is, the
more dissimilar the object features are. The dissimilarity measurement results are illustrated in Figure~\ref{fig:matching}. The OOD objects in the reference frame, such as street lights and billboards, have a more significant dissimilarity.%, which are effective alternatives 

% for an instance $h(\mathbf{x_0}, b_i)$, we find a matching target in frame $\mathbf{x}_1$, which should be as dissimilar as possible to become a proper unknown object in the feature space. 

% The blue arrow denotes the object that is the farthest from the object (in red) at time stamp $t_0$.

% \noindent\textbf{Find the unknown features.} 
% To determine the matched unknown objects
% across frames, 
% To control the diversity of the identified unknown instances, we follow the practice in the dense matching literature~\cite{DBLP:journals/corr/abs-2104-08381,DBLP:conf/cvpr/DwibediATSZ19,DBLP:conf/cvpr/PangQLCLDY21} and synthesize a soft OOD counterpart for each object, which is
Lastly, we perform a weighted average of the object features from frame $\mathbf{x}_1$. Using multiple objects captures a diverse distribution of unknowns. The weights $\alpha$ are defined as the normalized exponential of
the dissimilarity scores:
\vspace{-0.8em}
\begin{equation}
    \hat{\mathbf{o}}^{i}=\sum_{j=1}^{N_1} \alpha_{i, j} h(\mathbf{x_1}, \*b_j), \quad \alpha_{i, j}=\frac{e^{s_{i, j}}}{\sum_{k=1}^{N_1} e^{s_{i, k}}},
    \vspace{-0.8em}
    \label{eq:normalization}
\end{equation}
where $\hat{\mathbf{o}}^{i}$ is the distilled unknown object (in the feature space), corresponding to the $i$-th object at frame $\mathbf{x}_0$. %Note that previous approach synthesize images in the high-dimensional {pixel space}~\cite{lee2018training}, which can be intractable to optimize. Our unknown instances in the \emph{feature space}, which is more tractable given lower dimensionality.

% to avoid the
% instability caused by matching to a single instance feature. 

\vspace{-0.4cm}
\paragraph{Temporal unknown distillation.}~Our spatial unknown distillation mechanism operates on a single reference frame, which can be extended to multiple video frames to capture  additional diversity of unknowns in the \emph{temporal} dimension. For example, consider a video of a car driving on the highway, the more frames we observe, the more unknown objects can be observed, such as trees, buildings, and rocks. %Specifically, we now propose to synthesize the unknowns from objects across multiple frames. 

Given a frame $\*x_0$ at timestamp $t_0$, we propose distilling the unknown objects from multiple  frames $\mathbf{x}_1,...,\mathbf{x}_{T}$. We randomly sample $T$ frames within a range $[t_0-R, t_0+R]$. As a special case, $T=1$ reduces to the previous pair-frame setting.~To distill spatial-temporal unknown objects, we concatenate the object feature vectors from $T$ frames, and then measure their dissimilarity \emph{w.r.t} the objects in frame $\mathbf{x}_0$ by Equation~\eqref{eq:distance}. For the $i$-th object in frame $\mathbf{x}_0$, the unknown counterpart is defined as follows:
\vspace{-0.4em}
\begin{equation}
     \hat{\mathbf{o}}^{i}=    \sum_{j=1}^{N} \alpha_{i, j} h(\mathbf{x}, \*b_j), \*x\in \{\*x_1,...,\*x_T\},
     \label{eq:synthesis}
     \vspace{-0.4em}
\end{equation}
where $\alpha_{i,j}$ denotes the normalized dissimilarity scores defined in Equation~\ref{eq:normalization}. $N=\sum_{k=1}^{T} N_k$ is the total number of  objects across $T$ reference frames. The temporal aggregation mechanism allows searching through multiple frames for meaningful and diverse unknown discovery.%, which does not require refactoring the model architecture for training with the unknown-aware training objective.
%$\alpha_{i,j}^{k}$ is the dissimilarity between the $i$-th object in frame $\mathbf{x}_0$ and the objects in the $k$-th reference frame.

%and (2) sampling $T$ frames with a fixed interval $I$ where the sampled frames are at time stamps $t_0+kI,1 \leq k \leq T $. 
Later in Section~\ref{sec:ablation}, we provide comprehensive ablation studies on the frame sampling range $R$ and the number of selected frames $T$, and show the benefits of temporal aggregation for improved OOD detection.

\vspace{-0.2cm}
\paragraph{Unknown candidate object selection.}~A critical step in unknown distillation is to filter unknowns in the reference frame $\mathbf{x}_1$ that may be ID objects or simple background. Without selection, the model may be confused to separate the distilled unknown objects from the ID objects or quickly memorize the simple OOD pattern during training.  To prevent this, we pre-filter the proposals based on the energy score, and then use the selected ones for the spatial-temporal unknown distillation. It is shown that the energy score is an effective indicator of OOD data in image classification~\cite{liu2020energy}. To calculate the energy score for object detection network, we feed the object features $\{h(\mathbf{x}_1, \mathbf{b}_j)\}_{j=1}^{N_1}$ to the prediction head and follow the definition:
\vspace{-0.6em}
\begin{equation}
    E(\mathbf{x}_1, \mathbf{b}_j)=-\log \sum_{k=1}^{K} \exp ^{f_{k}\left(h(\mathbf{x}_1, \mathbf{b}_j); \boldsymbol{w}_{\text{pred}}\right)},
    \vspace{-0.5em}
    \label{eq:energy}
\end{equation}
where $f_{k}\left(h(\mathbf{x}_1, \mathbf{b}_j); \boldsymbol{w}_{\text{pred}}\right)$ is the logit output of the $k$-way classification branch. A higher energy indicates more OOD-ness and vice versa. Then, we select objects with mild energy scores, \emph{i.e.}, those in a specific percentile $p\% \leq\mathrm{Rank}(E(\mathbf{x}_1, \mathbf{b}_j)) / N_1\leq q\%$ among all objects. In case of multiple frames $\*x_1, \*x_2,...,\*x_T$, the object selection is performed on each individual frame before temporal aggregation. Ablation study on the effect of the energy filtering and the  selection percentile are provided in Section~\ref{sec:ablation}.

%to further select a proper set of unknown objects from frame $\mathbf{x}_1$. 

% With the \textit{temporal coherence} property of the videos, the proposals of the same object across frame $\mathbf{x}_0$ and $\mathbf{x}_1$ provide us positive matching pairs where the distance 

\subsection{Unknown-Aware Training Objective}
\label{sec:regularization}
Leveraging the distilled unknown objects from Section~\ref{sec:unknown_distill}, we now introduce our training objective for unknown-aware object detection.  Our key idea is to perform object detection task while regularizing the model to produce a low uncertainty score for ID objects,  and a high uncertainty score for the unknown ones. The overall objective function is defined as:
\vspace{-0.5em}
\begin{equation}
   \mathcal{L} =  \mathcal{L}_\text{det} +\beta \cdot \mathcal{L}_\text{uncertainty},
   \vspace{-0.5em}
   \label{eq:all_loss}
\end{equation}
where $\beta$ is the scaling weight when combining the detection
loss $\mathcal{L}_\text{det}$ and the uncertainty regularization loss $\mathcal{L}_\text{uncertainty}$. Next we describe the details of  $\mathcal{L}_\text{uncertainty}$. 
\vspace{-0.7em}

% \noindent\textbf{Uncertainty score.} 

\begin{figure}[t]
    \centering
    \includegraphics[width=1.0\linewidth]{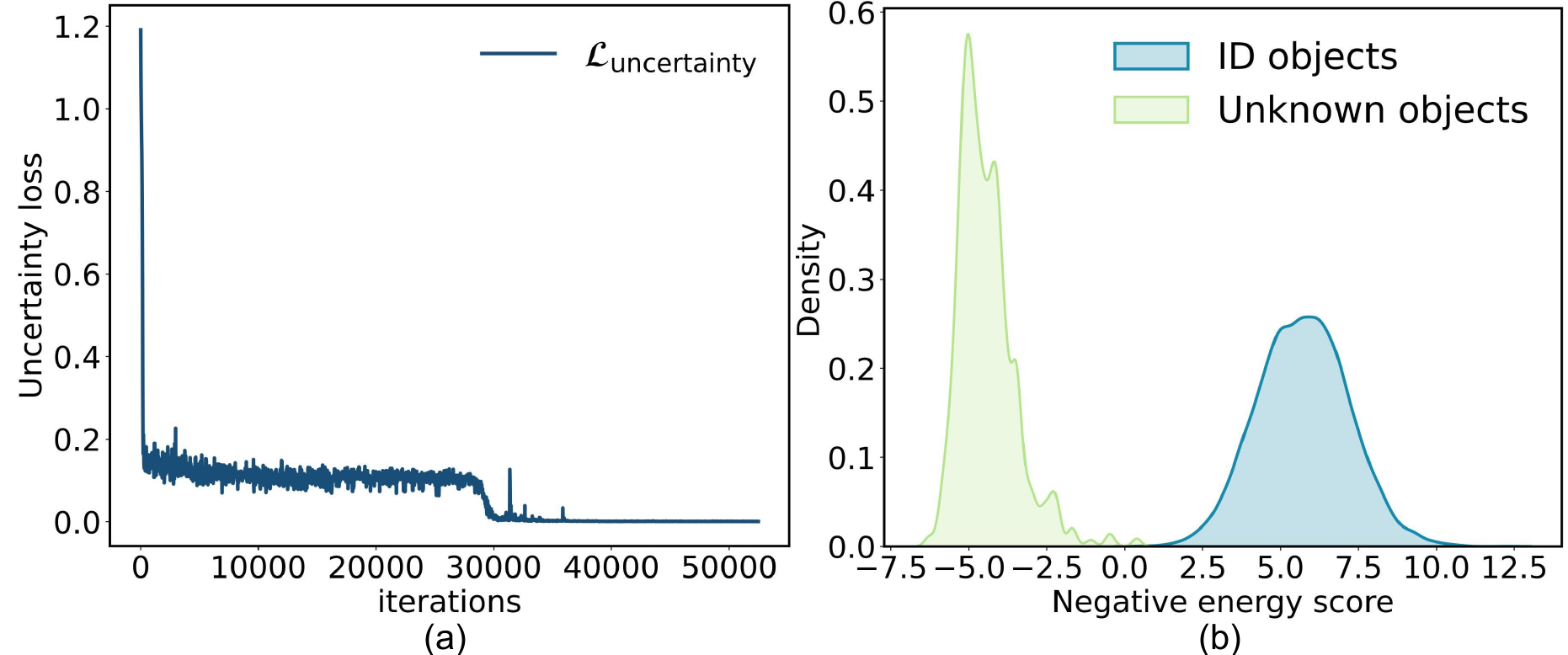}
      \vspace{-2em}
    \caption{\small (a) Uncertainty regularization loss during training. (b) The negative energy score distribution for both the ID and the distilled unknown objects after training.   }
    \label{fig:convergence}
    \vspace{-1em}
\end{figure}

\vspace{-0.3cm}
\paragraph{Uncertainty regularization.} 
Following Du \emph{et al.}~\cite{du2022vos}, we employ a loss function that contrastively shapes the uncertainty surface, amplifying the separability between known ID objects and  unknown OOD objects. 
To measure the uncertainty, we use the energy score in Equation~\eqref{eq:energy}, which is derived from the output of the classification branch. Here we calculate the energy score $E(\mathbf{x},\mathbf{b})$ for the ID objects and the distilled unknown object features $E(\hat{\mathbf{o}})$. The uncertainty score is then passed into a logistic regression classifier with weight coefficient $\theta_{\mathrm{u}}$, which predicts high probability for ID object $(\mathbf{x}, \mathbf{b})$ and low probability for the unknown ones $\hat{\mathbf{o}}$. The regularization loss is calculated as:
\begin{equation}
\small
    % \vspace{-0.7em}
\begin{aligned}
     \mathcal{L}_{\text {uncertainty }}&=\mathbb{E}_{\hat{\mathbf{o}} \sim \mathcal{O}}\left[-\log \frac{1}{1+\exp ^{-\theta_{\mathrm{u}} \cdot E(\hat{\mathbf{o}})}}\right]+\\& \mathbb{E}_{(\mathbf{x}, \mathbf{b}) \sim \mathcal{D}}\left[-\log \frac{\exp ^{-\theta_{\mathrm{u}} \cdot E(\mathbf{x}, \mathbf{b})}}{1+\exp ^{-\theta_{\mathrm{u}} \cdot E(\mathbf{x}, \mathbf{b})}}\right],
     \label{eq:reg_loss}
\end{aligned}
    % \vspace{-0.5em}
\end{equation}
where $\mathcal{O}$ contains all the unknown object features (\emph{c.f.} Section~\ref{sec:unknown_distill}). In Figure~\ref{fig:convergence}(a), we show the uncertainty regularization loss $\mathcal{L}_\text{uncertainty}$ over the course of training on Youtube-VIS dataset~\cite{DBLP:conf/iccv/YangFX19}. Upon convergence, Figure~\ref{fig:convergence}(b) shows the energy score distribution for both the ID and distilled unknown objects. This demonstrates that \texttt{\model} converges properly and is able to separate the distilled unknown objects and the ID objects.

% \begin{figure}[t]
%     \centering
%     \includegraphics[width=1.0\linewidth]{./figs/convergence_plot.pdf}
%     \caption{\small Visualization of the (a) uncertainty regularization loss during training and (b) the negative energy score distribution for both the ID and the distilled unknown instances.   }
%     \label{fig:convergence}
%     \vspace{-1em}
% \end{figure}

\begin{algorithm}[H]

\SetAlgoLined
\textbf{Input:} ID data $\mathcal{D}=\left\{\left(\*x_{i}, \mathbf{b}_i,{y}_{i}\right)\right\}_{i=1}^{M}$, randomly initialized object detector with parameter $\theta$, energy filtering percentile $[p\%,q\%]$, sampling range $R$, the number of reference frames $T$, and weight for uncertainty regularization $\beta$.\\

\textbf{Output:} Object detector with parameter $\theta^{*}$, and OOD detector $G$.\\
\While{train}{

1. Select unknown objects in the reference frames  with mild energy scores as defined by Equation~\eqref{eq:energy}.

2. Calculate the dissimilarity (using Equation~\eqref{eq:normalization}) between an object in the key frame \emph{w.r.t} selected objects in the reference frames.

3. Distill the unknown objects by Equation~\eqref{eq:synthesis}.
  
  4. Calculate the uncertainty regularization loss by Equation~\eqref{eq:reg_loss}, update the parameter $\theta$ based on the total loss in Equation~\eqref{eq:all_loss}.
  }
 
  \While{eval}{
  1.~Calculate the uncertainty score by Equation~\eqref{eq:ood_uncertainty}.\;
  
  2.~Perform thresholding comparison by Equation~\eqref{eq:ood_detection}.

 }
 \caption{\model: Spatial-Temporal Unknown Distillation for OOD detection}
 
 \label{alg:algo}
\end{algorithm}
\vspace{-1em}
Compared to $\mathcal{L}_{\text{det}}$ for the vanilla object detector, our loss intends to facilitate learning a more conservative decision boundary between ID and OOD objects, which helps flag unseen OOD objects in testing. 
 We proceed by describing the test-time OOD detection procedure.
 
% The learning process allows greater flexibility in contrastively shaping the uncertainty surface, resulting in more distinguishable in- and out-of-distribution objects.
% ~\cite{liu2020energy} employed energy for model uncertainty regularization, however, the loss function is based on the squared hinge loss and requires tuning two margin hyperparameters. In contrast, our uncertainty regularization loss is completely \emph{hyperparameter-free} and is much easier to use in practice. Moreover, \model produces a probabilistic score for OOD detection, whereas \cite{liu2020energy} relies on a non-probabilistic energy score. We provide ablation studies in Section \model demonstrating the superiority of our loss function.

\vspace{-0.4cm}
\paragraph{Test-time OOD detection.}
\label{sec:inference}
During inference, we use the output of the logistic regression uncertainty branch for OOD detection. In particular, given a test input $\mathbf{x}^*$, the object detector produces a box prediction $\mathbf{b}^*$. The uncertainty score for the predicted object $(\mathbf{x}^*, \mathbf{b}^*)$ is given by:
\vspace{-0.3em}
\begin{align}
    p_\theta(g \mid \mathbf{x}^*, \mathbf{b}^*) = \frac{\exp^{- {\theta}_{\mathrm{u}}\cdot E(\mathbf{x}^*, \mathbf{b}^*)}}{1+\exp^{- {\theta}_{\mathrm{u}}\cdot E(\mathbf{x}^*, \mathbf{b}^*)}}.
    \vspace{-2.9em}
    \label{eq:ood_uncertainty}
\end{align}
% \begin{equation}
%     G(\hat{b}):=\left\{\begin{array}{ll}
% 1 & \text { if }-E(\hat{b}) \geq \gamma, \\
% 0 & \text { if }-E(\hat{b})<\gamma,
% \end{array}\right.
% \label{eq:ood_detection}
% \end{equation}

For OOD detection, we use the common thresholding mechanism to  distinguish
between ID and OOD objects:
\vspace{-0.2em}
\begin{equation}
    G(\mathbf{x}^*, \mathbf{b}^*)=\left\{\begin{array}{ll}
1 & \text { if }p_\theta(g \mid \mathbf{x}^*, \mathbf{b}^*)\geq \gamma, \\
0 & \text { if }p_\theta(g \mid \mathbf{x}^*, \mathbf{b}^*) <\gamma.
\end{array}\right.
\vspace{-0.2em}
\label{eq:ood_detection}
\end{equation}
The threshold $\gamma$ is typically chosen so that a high
fraction of ID data (e.g., 95\%) is correctly classified. For objects that are classified as ID, one can obtain the bounding box and class prediction using the prediction head as usual. Our approach \model is summarized in Algorithm~\ref{alg:algo}.

% \vspace{-1.9cm}

\vspace{-1.3em}
\paragraph{Synergy between unknown distillation and contrastive regularization.} The two key components of \texttt{\model}---unknown distillation (Section~\ref{sec:unknown_distill}) and contrastive regularization (Section~\ref{sec:regularization}) work collaboratively.
First, a set of well distilled unknown objects may improve the energy-based contrastive regularization and help learn a more accurate decision boundary between known and unknown objects. Second, as the contrastive uncertainty loss amplifies an energy gap between known and unknown objects, the unknown distillation module can benefit from more accurate unknown object selection (via energy-based filtering). 
The entire training process converges when the two components perform satisfactorily.
Our experiments in Section~\ref{sec:experiments} further justify the efficacy of our framework.

% \vspace{-1em}

% \vspace{-0.8em}

\section{Experiments}
\label{sec:experiments}

In this section, we provide empirical evidence to show the effectiveness of \texttt{\model} on two large-scale video datasets (Section~\ref{sec:setting}). We show that \texttt{\model} outperforms other commonly used OOD detection baselines on detecting OOD data in Section~\ref{sec:baselines}. Ablation studies of \texttt{\model}  and qualitative analysis are presented in Sections~\ref{sec:ablation} and~\ref{sec:visualization}.  

%\scalebox{0.8}{

\begin{table*}[t]
\centering
\scalebox{0.8}{\begin{tabular}{llllll}
    \toprule
    {In-distribution $\mathcal{D}$}  & 
   {\textbf{Method }}  &\textbf{FPR95} $\downarrow$  & \textbf{AUROC} $\uparrow$ & \textbf{ mAP (ID)}$\uparrow$ & \textbf{Cost (h)} \\ \hline
    && \multicolumn{2}{c}{OOD: MS-COCO / nuImages}\\
     \cline{3-4}
    \multirow{9}{0.2\linewidth}{{\textbf{ BDD100K} }} 
    & MSP \cite{hendrycks2016baseline}
   & 90.11 / 93.98 & 66.32 / 59.21&31.0 &9.1 \\
     & ODIN \cite{liang2018enhancing}
    & 80.32 / 87.75 &  68.49 / 66.51&31.0 &9.1 \\
     & Mahalanobis 
    \cite{lee2018simple}
    & 63.06 / 79.02 &  79.95 / 68.94 & 31.0 &9.1\\ 
        & Gram matrices~\cite{DBLP:conf/icml/SastryO20}&68.78  / 82.60 &66.13 / 71.56 	&31.0  &9.1\\
      & Energy score~\cite{liu2020energy}
    & 78.36 / 86.02 &  73.75 / 67.08& 31.0 &9.1\\ 

     & Generalized ODIN~\cite{hsu2020generalized} & 75.99 / 92.15  & 78.63 / 67.23&30.9&10.5
     \\
  & CSI~\cite{tack2020csi}&69.38 / 80.06 & 80.85 /  72.59  &29.8&15.3\\
    & GAN-synthesis~\cite{lee2018training}&	67.95 / 88.53 & 78.33 / 66.50&30.1 &14.6\\
        & \cellcolor{Gray}\textbf{\model}  (ours)  &\cellcolor{Gray}\textbf{52.18}$\pm$2.2 / \cellcolor{Gray}\textbf{77.57}$\pm$3.0 & \cellcolor{Gray}\textbf{85.67}$\pm$0.6 / \cellcolor{Gray}\textbf{75.67}$\pm$0.7&
\cellcolor{Gray}30.5$\pm$0.2&\cellcolor{Gray}10.1 \\
     \midrule
 \multirow{9}{0.2\linewidth}{\textbf{Youtube-VIS} } 
     & MSP \cite{hendrycks2016baseline}
    & 90.17 / 94.52 &  70.26 / 54.59	& 24.8  &9.2\\
     & ODIN \cite{liang2018enhancing}
    &87.17 / 97.69 & 71.46 / 57.46 & 24.8  &9.2\\
     & Mahalanobis 
    \cite{lee2018simple}
   	& 85.60 / 95.65 & 72.16 / 62.02& 24.8 &9.2\\ 
   	    & Gram matrices~\cite{DBLP:conf/icml/SastryO20}&	88.68 / 93.20& 61.96 / 60.04 	&24.8  &9.2 \\
      & Energy score~\cite{liu2020energy}
    &91.77 / 91.78   & 70.58 / 59.05& 24.8 &9.2\\ 

      &Generalized ODIN~\cite{hsu2020generalized} & 83.90 / 93.18 &  71.33 / 62.16&24.3&10.5
     \\
     &  CSI~\cite{tack2020csi}&80.21 / 84.85 &   73.89 / 68.84& 23.3&15.7 \\
    & GAN-synthesis~\cite{lee2018training}&	84.57  / 94.59 &  71.59 / 64.43 & 24.4&15.0\\
     & \cellcolor{Gray}\textbf{\model}  (ours) 
    & 	\cellcolor{Gray}\textbf{79.82}$\pm$0.2 / \textbf{76.93}$\pm$0.4 &	\cellcolor{Gray}\textbf{75.55}$\pm$0.3 / \textbf{71.48}$\pm$0.6	& \cellcolor{Gray}24.5$\pm$0.3 &\cellcolor{Gray}10.2\\
        \bottomrule
\end{tabular}}
        \vspace{-0.2cm}
        \caption[]{\small \textbf{Main results.} Comparison with competitive out-of-distribution detection methods. All baseline methods are based on a model trained on \textbf{ID data only} using ResNet-50 as the backbone.  $\uparrow$ indicates larger values are better, and $\downarrow$ indicates smaller values are better. All values are percentages. \textbf{Bold} numbers are superior results.  We report standard deviations estimated across three runs. The training time is reported in the ``cost" column on four NVIDIA GeForce RTX 2080Ti GPUs. }
        \label{tab:baseline}
        \vspace{-0.8em}
\end{table*}

\subsection{Benchmark Construction}
\label{sec:setting}
\noindent\textbf{Datasets.}  We use two large-scale video datasets as ID  data: \texttt{BDD100K}~\cite{DBLP:conf/cvpr/YuCWXCLMD20} and Youtube-Video Instance Segmentation (\texttt{Youtube-VIS}) 2021~\cite{DBLP:conf/iccv/YangFX19}. For both tasks, we evaluate on two OOD datasets containing diverse visual categories: \texttt{MS-COCO}~\cite{lin2014microsoft} and \texttt{nuImages}~\cite{DBLP:conf/cvpr/CaesarBLVLXKPBB20}.  We perform careful deduplication to ensure there is no semantic overlap between the ID and OOD data. Extensive details on the datasets are described in the appendix. 
\vspace{-0.4cm}
\paragraph{Implementation details.} We adopt Faster R-CNN~\cite{ren2015faster} as the base object detector. We use Detectron2 library~\cite{Detectron2018} and train with the backbone of ResNet-50~\cite{DBLP:conf/cvpr/HeZRS16} and the default hyperparameters. We set the weight $\beta$ for $\mathcal{L}_{\text{uncertainty}}$ to be $0.05$ for BDD100K and $0.02$ for Youtube-VIS dataset. For both datasets, we use $T=3$ frames and set the sampling range $R=9$. We set the energy filtering percentile to be $40\%-60\%$ among all proposals. Ablation studies on different hyperparameters are detailed in Section~\ref{sec:ablation}. 

%we set the frame range $R$ and frame interval $I$ to be 7 and 5 for pair-frame setting. For multi-frame setting,

\vspace{-0.4cm}
\paragraph{Metrics.} For evaluating the OOD detection performance, we report: (1) the false positive rate (FPR95) of
OOD samples when the true positive rate of ID
samples is at 95\%; (2) the area under the receiver operating characteristic curve (AUROC). For evaluating the object detection performance on the ID task, we report the common metric of mAP.

\subsection{Comparison with Baselines}
\label{sec:baselines}

\paragraph{\model establishes SOTA performance.} In Table~\ref{tab:baseline}, we compare \texttt{\model} with competitive OOD detection methods in literature, where \texttt{\model} significantly outperforms baselines on both datasets. For a fair comparison, all the methods use the same ID training data, trained with the same number of epochs.  Our comprehensive baselines include Maximum Softmax Probability~\cite{hendrycks2016baseline}, ODIN~\cite{liang2018enhancing}, Mahalanobis distance~\cite{lee2018simple}, Generalized ODIN~\cite{hsu2020generalized}, energy score~\cite{liu2020energy},\@
%entropy score~\cite{DBLP:journals/corr/abs-2101-05036}, 
Gram matrices~\cite{DBLP:conf/icml/SastryO20}, and a latest method CSI~\cite{tack2020csi}. These baselines rely on the classification output or backbone feature, and therefore can be seamlessly evaluated on the object detection model. 

The results show that \model can outperform these baselines by a considerable margin because the majority of baselines rely on object detection models trained
on ID data only, without being regularized by unknown objects. Such a training scheme is prone to produce overconfident predictions on OOD data (Figure~\ref{fig:teaser}) while \model  incorporates unknown objects to regularize the model more
effectively.

 We also compare with GAN-based approach for synthesizing outliers in the pixel space~\cite{lee2018training}, where \texttt{\model} effectively improves the OOD detection performance (FPR95) by \textbf{15.77}\% on BDD100K (COCO as OOD) and \textbf{17.66}\% on Youtube-VIS (nuImages as OOD). Moreover, we show in Table~\ref{tab:baseline} that \texttt{\model} achieves stronger OOD detection performance while preserving a high object detection accuracy on ID data (measured by mAP). This is in contrast with CSI, which displays significant degradation, with mAP decreasing by 1.2\% on Youtube-VIS. \emph{Details of reproducing baselines are in the Appendix Section~\ref{sec:reproduce_baseline}}.

%Finally, we demonstrate that distilling diverse unknowns by aggregating multiple frames is beneficial for OOD detection, i.e., improving the FPR95 by \textbf{9.23}\% on BDD-100k (COCO as OOD).

% Highlight the best results under the two-frame and multi-frame setting. Each of them can be implemented by random sampling within a range and sampling with a fixed frame interval. Results on a different backbone, etc. 

% Baselines are 

%w/ Pair-frame means we select $T=1$ adjacent frame for unknown distillation while w/ Multi-frame denotes $T=3$

% \subsection{Results on Youtube-VIS dataset}
% Highlight the best results under the two-frame and multi-frame setting. Each of them can be implemented by random sampling within a range and sampling with a fixed frame interval. Results on a different backbone, etc.

% Baselines are Maximum Softmax Probability~\cite{hendrycks2016baseline}, ODIN~\cite{liang2018enhancing}, Mahalanobis distance~\cite{lee2018simple}, Generalized ODIN~\cite{hsu2020generalized}, energy score~\cite{liu2020energy},\@
% %entropy score~\cite{DBLP:journals/corr/abs-2101-05036}, 
% CSI~\cite{tack2020csi} and Gram matrices~\cite{DBLP:conf/icml/SastryO20} and GAN-based approach for synthesizing outliers~\cite{lee2018training}

\subsection{Ablation Studies}
\label{sec:ablation}

% \noindent\textbf{Convergence visualization.} In Figure~\ref{fig:convergence}, we show the trend of the uncertainty regularization loss during training and the energy score distribution for both the ID and distilled unknown instances. We use the Youtube-VIS as the ID training dataset. The results demonstrate that \texttt{\model}  converges stably, and can separate the distilled unknown instances and the ID instances well.

\begin{figure*}[t]
    \centering
    \begin{subfigure}{.45\textwidth}
  \centering
%   \vspace{-0.5em}
  \includegraphics[width=1.0\linewidth]{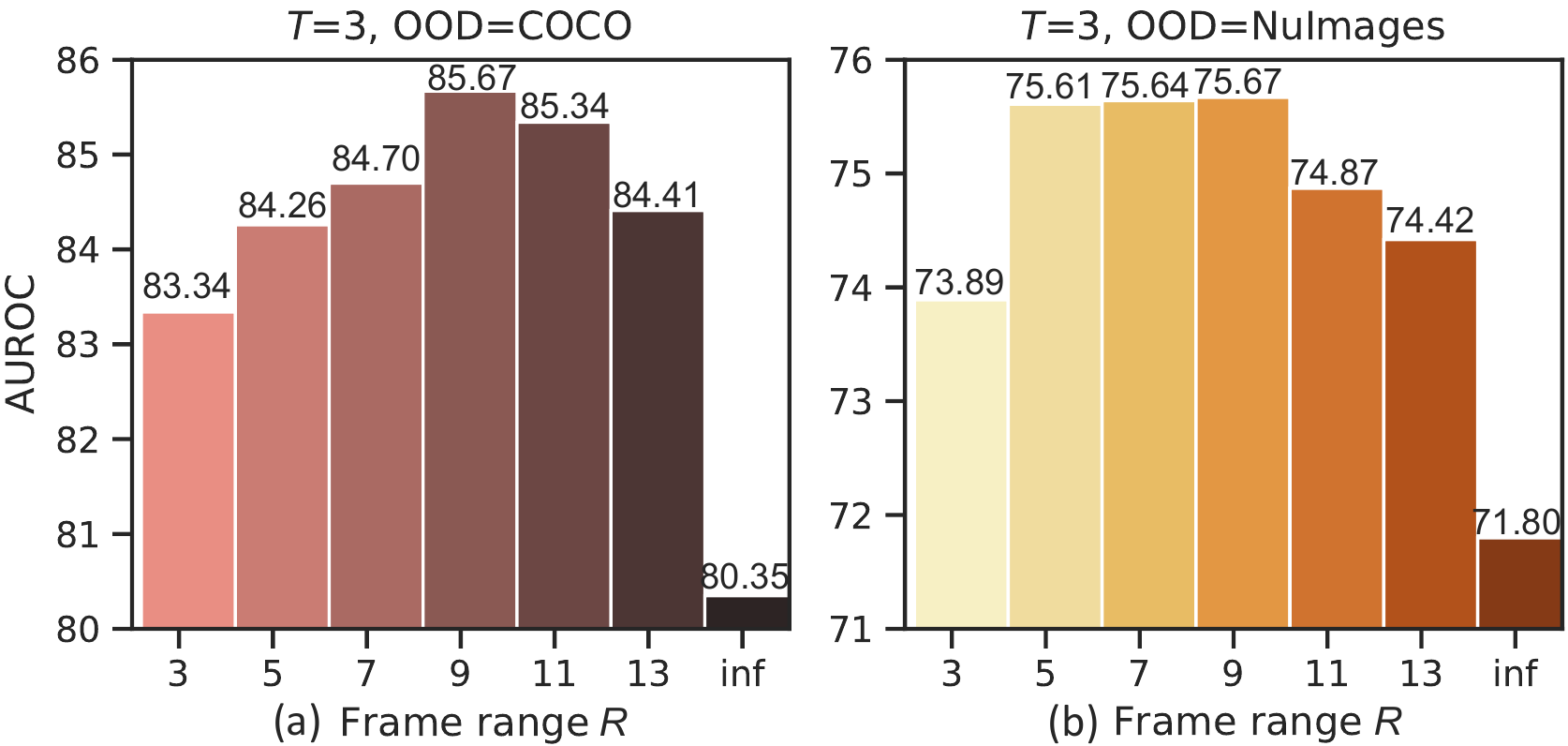}
%   \caption{A subfigure}
  \label{fig:sub1}
\end{subfigure}
 \begin{subfigure}{.45\textwidth}
  \centering
  \includegraphics[width=1.0\linewidth]{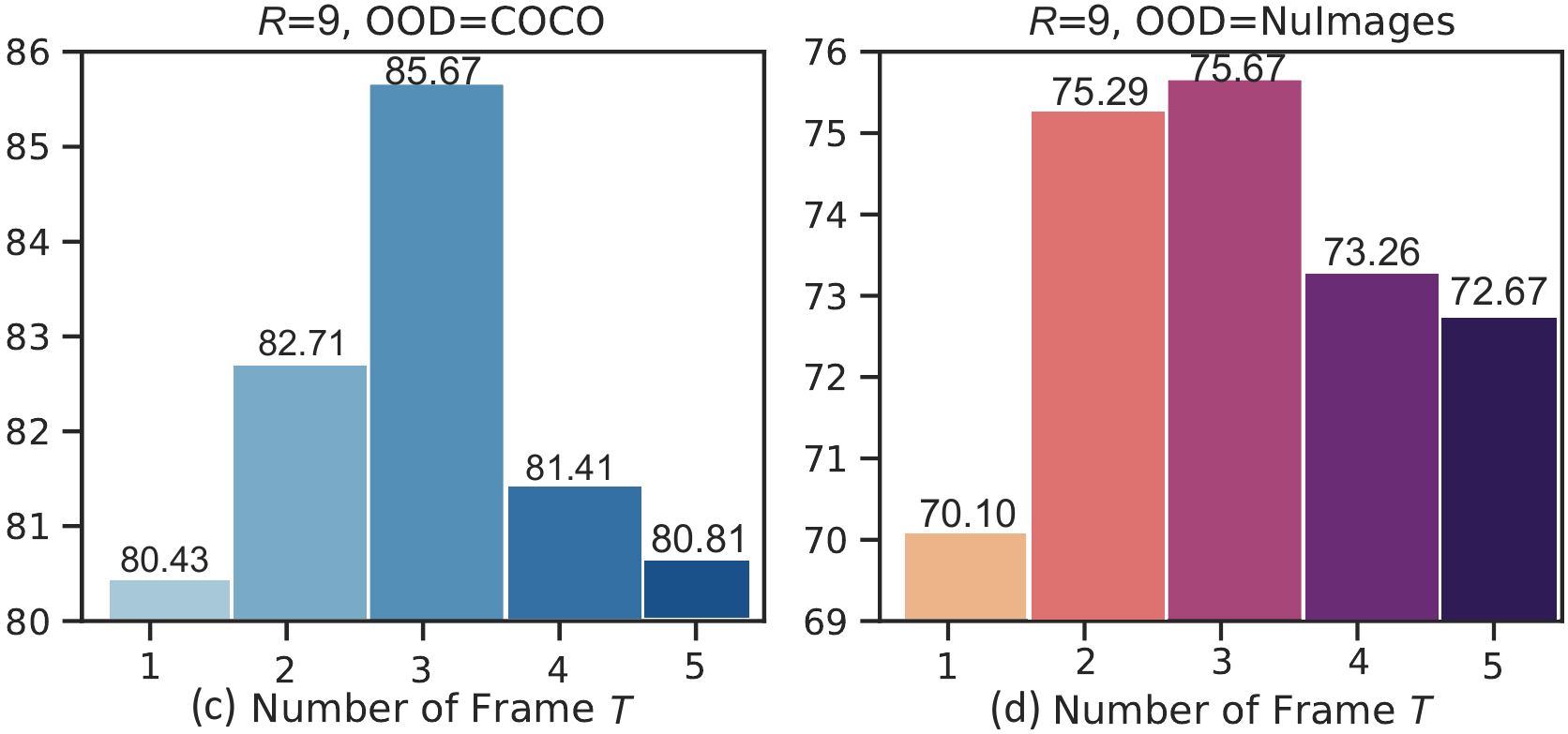}
%   \caption{A subfigure}
  \label{fig:sub2}
\end{subfigure}
    \vspace{-2em}
    \caption{\small (a)-(b) Ablation study on the  sampling range $R$. We vary the range from 3 to infinity. Metrics are AUROC. We set $T=3$. (c)-(d) Ablation study on the number of reference frames $T$ during unknown distillation. We fix the sampling range as $R=9$. }
    \label{fig:ablation_frame_range}
    \vspace{-1em}
\end{figure*}

This section provides comprehensive ablation studies to understand the efficacy of \model. For consistency, all ablations are conducted on the BDD100K dataset, using ResNet-50 as the backbone. We refer readers to Appendix Section~\ref{sec:backbone_app} for more ablations on using a different backbone architecture.

\begin{table}[t]
\centering
\scalebox{0.85}{
\begin{tabular}{lcc}
    \toprule \multirow{2}{0.08\linewidth}{\textbf{Method }}   & \textbf{AUROC} $\uparrow$ & \textbf{mAP}  $\uparrow$ \\
    \hline
& COCO / nuImages as OOD\\
    \cline{2-2} 
    $^\diamond$Farthest object &  83.04 / 71.38& 30.2\\
      $^\diamond$Random object &     79.61 / 70.42& 30.3\\
        $^\diamond$Object with mild energy&     83.60 / 71.24& 30.3\\
          $^\diamond$Negative proposal~\cite{DBLP:journals/corr/abs-2103-02603} &     80.94 / 72.92& 30.0\\
    \hline
    
    % $^\natural$ xxx w/ infinite frame range&    80.35 / 71.80&30.6\\
    % \hline
  $^\clubsuit$GAN~\cite{lee2018training} &          78.33 / 66.50 & 30.1 \\
  $^\clubsuit$Mixup~\cite{DBLP:conf/iclr/ZhangCDL18} &    81.76 / 70.17 & 27.6 \\
 \midrule

  $^\natural$Gaussian noise & 83.64 / 71.50 & 30.3\\
 \midrule 
    \rowcolor{Gray} \textbf{\model}   (ours) 
    &	\textbf{85.67} / \textbf{75.67}	& 30.5\\
        \bottomrule
\end{tabular}}
        \vspace{-0.2cm}
        \caption[]{\small {Ablation on different unknown distillation approaches (on backbone of ResNet-50, COCO / nuImages are the OOD data). } }
        \label{tab:synthesis_ablation}
        \vspace{-0.5cm}
\end{table}

\vspace{-0.45cm}
\paragraph{Ablation on different unknown distillation approaches.} We compare  \texttt{\model} with three types of unknown distillation approaches, \emph{i.e.}, (I$^\diamond$) using independent objects without spatial-temporal aggregation, %(ii$^\natural$) synthesis w/o relying on the key frame, 
(II$^\clubsuit$) synthesizing unknowns in the pixel space, and (III$^\natural$) using noise as unknowns. %All methods are trained under the same setup, with BDD-100k as ID data and ResNet-50 as the backbone.
\begin{itemize}[leftmargin=1em]
\vspace{-0.5em}
    \item For \textbf{type I}, we utilize objects from the reference frame without aggregating multiple objects across spatial and temporal dimensions---a key difference from \texttt{\model}. The unknown objects can be constructed by: using the object in the reference frame that has the largest dissimilarity, using random objects, using the negative object as in~\cite{DBLP:journals/corr/abs-2103-02603}, and using objects with mild energy scores (percentile $40\%-60\%$) in the reference frame.
    \vspace{-0.7em}
    \item For \textbf{type II}, we consider GAN-based~\cite{lee2018training} and mixup-based~\cite{DBLP:conf/iclr/ZhangCDL18} methods. For~\cite{lee2018training}, the classification outputs of the objects in the synthesized images are forced to be closer to a uniform distribution.~For mixup, we use a beta distribution of $\operatorname{Beta}(1)$, and interpolate ID objects in the pixel space for the reference frames. 
    %  \vspace{-1.5em}
        \vspace{-0.7em}
    \item For \textbf{type III}, we add fixed Gaussian noise to the ID objects to create unknown object features. 
    % \vspace{-0.5em}
\end{itemize}

%For \textbf{type II}, we set the frame range to infinity so that \texttt{\model} reduces to randomly sampling two images from the entire dataset rather than requiring a key frame. 

The results are summarized in Table~\ref{tab:synthesis_ablation}, where  \texttt{\model} outperforms alternative approaches. Exploiting objects without spatial-temporal distillation ($\diamond$) is less effective than  \texttt{\model}, because the generated unknowns either lack diversity (\emph{e.g.,} using object with the biggest dissimilarity or mild energy) or are too simple to effectively regularize the decision boundary between ID and OOD (\emph{e.g.,} using negative or random objects). Synthesizing unknowns in the pixel space ($\clubsuit$) is either unstable (GAN) or harmful for the object detection performance (mixup). Lastly, Gaussian noise as unknowns is relatively simple and does not outperform  \texttt{\model}.

\vspace{-0.35cm}
\paragraph{Ablation on candidate object selection.} Table~\ref{tab:energy_filtering} investigates the importance of filtering unknown objects based on the energy score. We contrast performance by either removing the filtering, or using different filtering percentile (\emph{c.f.}  Section~\ref{sec:unknown_distill}). Using the objects with a mild energy score in the reference frames performs the best. This strategy distills unknown objects with a proper difficulty level, which is effective during contrastive uncertainty regularization.

\begin{table}[t]
    \centering
    \footnotesize
    \begin{tabular}{c|ccc}
    \hline
        Variants & FPR95 $\downarrow$  & AUROC $\uparrow$  &mAP $\uparrow$   \\
        \hline
    
        & \multicolumn{2}{c}{\small COCO / nuImages as OOD}\\
        \cline{2-3}
         \textbf{w/o} unknown filtering  &  62.23 / 83.54 &82.87 /  72.29&30.6 \\
      \textbf{w/}  ratio 0\%-20\%   &  61.41 / 82.33 &   83.66 / 74.86 &30.2 \\
      \textbf{w/}  ratio 20\%-40\% & 57.73 / 82.13 & 85.43 / 74.09  & 30.3 \\
       \textbf{w/}  ratio 40\%-60\% & \textbf{52.18} / \textbf{77.57}& \textbf{85.67} / \textbf{75.67}&
     30.5 \\
        \textbf{w/}  ratio  60\%-80\%& 62.29 / 85.12 & 83.47 /  73.44&30.2\\
     \textbf{w/}  ratio 80\%-100\%& 65.86 / 88.47 &  82.46 / 72.50 &30.3\\
         \hline
    \end{tabular}
    \vspace{-1em}
    \caption{\small Ablation study on the energy filtering module.  Here we set $T=3$ and $R=9$.}
    \label{tab:energy_filtering}
    \vspace{-2em}
\end{table} 
%Here we randomly sample 3 frames within a sampling range of 9.
% Comparison with 1) remove the energy filtering, 2) different selection percentage in energy filtering.

% Comparison with 1) choosing the farthest one, 2) the random one in a different frame, 3) use Gaussian noise, 4) mild energy proposals, 5) large frame range and interval , 6) negative proposals and 7) data augmentation (mixup).

\vspace{-0.35cm}
\paragraph{Ablation on the frame sampling range ${R}$.} Recall our spatial-temporal unknown distillation requires concatenation of objects from $T$ reference frames. We ablate the effect of randomly selecting $T$ frames within different temporal horizons \emph{w.r.t} the key frame, modulated by the sampling range $R$. 
%To trade off exploiting the video consistency property and exploring the unknown diversity for instance-level unknown distillation, we show the effect of the frame range $R$ if we randomly select $T$ reference frames within $R$. 
The results with varying $R$ are shown in Figure~\ref{fig:ablation_frame_range} (a)-(b) with $T=3$. We observe that OOD detection benefits from using the reference frames that are mildly close to the key frame. The trend is consistent for both COCO and nuImages OOD datasets. A larger sampling range translates into more dissimilar scenes, resulting in relatively easier unknowns to be distilled. When $R$ becomes infinity, {\model} {randomly samples frames from the entire video}, where the distilled unknowns are much less effective with AUROC significantly degrades (from 85.67\% to 80.35\% on COCO).% BDD-100k is used as the in-distribution dataset for training.

%incurs a completely different reference frame which is difficult for the dissimilarity measurement to identify and distill the unknown object instances.  

% The ablation assumes we use random sampling.

% Frame range is set to \{3,5,7,9,11,13,15\} for BDD dataset and \{3,5,7,9,10,11,13,15\} for VIS dataset.

\vspace{-0.35cm}
\paragraph{Ablation on the number of reference frames $T$.}  We contrast performance under different number of reference frames $T$ and report the OOD detection results in Figure~\ref{fig:ablation_frame_range} (c)-(d). This ablation shows that \model indeed benefits from aggregating objects from multiple frames across the temporal dimension. For example, the model trained on BDD100K with $T=3$ achieves an AUROC improvement of 5.24\%  (COCO as OOD) compared to $T=1$. This highlights the importance of temporal distillation with multiple frames. However, a larger $T$ hurts the OOD detection performance. We hypothesize this is because many redundant object features are used during unknown distillation. %Moreover, the mAP decreases because each training batch contains similar images that prevent the model from generalizing well. \SL{check claims}

\begin{figure*}[t]
    \centering
    \includegraphics[width=1.0\textwidth]{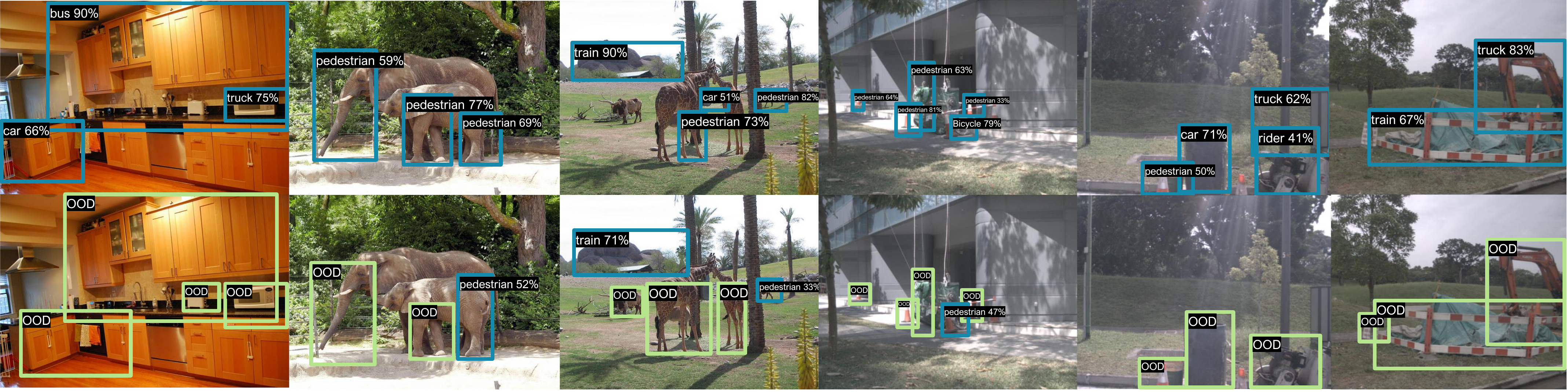}
    \vspace{-1.8em}
    \caption{\small Visualization of detected objects on the OOD images (from MS-COCO and nuImages) by a vanilla Faster-RCNN (\emph{top}) and \texttt{\model} (\emph{bottom}). The ID is BDD100K dataset. \textbf{Blue}: OOD objects classified as one of the ID classes. \textbf{Green}: OOD objects detected by \texttt{\model}, which reduce false positives among detected objects. Additional visualization is shown in Appendix Section~\ref{sec:app_visual}.}
% \vspace{-2em}
\vspace{-1.5em}
\label{fig:visual}
\end{figure*}

\vspace{-0.35cm}
\paragraph{Ablation on the uncertainty regularization weight $\beta$.} Table~\ref{tab:loss_weight} reports the OOD detection results as we vary the weight $\beta$ for $\mathcal{L}_\text{uncertainty}$. The model is evaluated on both COCO and nuImages datasets as OOD. %$\{0.03,0.04,0.05,0.06,0.07\}$ with the BDD-100k as the ID training data and evaluated on COCO and NuImages dataset in . 
%Here we randomly sample adjacent frames within a frame range of 9. 
The results suggest that a mild weight is desirable. In most cases, \model outperforms the baseline OOD detection methods in Table~\ref{tab:baseline} in terms of AUROC.%As expected, a large value (e.g., $\beta=0.07$) will over-regularize the model and harm the performance. 

\begin{table}[t]
    \centering
    \footnotesize
    % \small
\scalebox{1.14}{    \begin{tabular}{c|ccc}
    \hline
        $\beta$ & FPR95 $\downarrow$  & AUROC $\uparrow$  &mAP $\uparrow$   \\
        \hline
    
        & \multicolumn{2}{c}{\small COCO / nuImages as OOD}\\
        \cline{2-3}
         0.03 & 63.52 / 86.18 &  83.49 / 70.70 &  30.4\\
     0.04   &59.52 / 84.01  & 84.03 / 72.09   &30.3 \\
        0.05 & \textbf{52.18} / \textbf{77.57}& \textbf{85.67} / \textbf{75.67}&
    30.5\\
     0.06& 57.37 / 85.53  & 84.59 / 72.60&30.2\\
     0.07 & 55.03 / 84.43 &  84.18 / 71.21  &30.2\\

         \hline
    \end{tabular}}
        \vspace{-1em}
    \caption{\small Ablation study on the weight $\beta$ for the uncertainty regularization loss. In this case, we set $T=3$ and $R=9$. }
    \label{tab:loss_weight}
    \vspace{-2em}
\end{table}

% The regualarization weight is set to \{0.03,0.04,0.05,0.06,0.07\} for BDD dataset and \{0.005,0.01,0.02,0.03,0.04\} for VIS dataset.

\vspace{-0.35cm}
\paragraph{Ablation on the uncertainty loss.} We perform ablation on three alternatives for $\mathcal{L}_\text{uncertainty}$: (1) using the squared hinge loss~\cite{liu2020energy}, (2) classifying the unknowns as an additional $K+1$ class in the classification branch and (3) removing the weight $\theta_{\mathrm{u}}$ in $\mathcal{L}_\text{uncertainty}$. 
%; and (3) using the same $\mathcal{L}_\text{uncertainty}$ as in Equation~\ref{eq:reg_loss} but with different loss weight $\beta$. 
The comparison is summarized in Table~\ref{tab:ablation_loss}.  Compared to the hinge loss, our logistic loss improves the AUROC by 11.35\% (COCO as OOD). 
In addition, classifying the distilled unknowns as an additional class increases the difficulty of object classification, which does not outperform either. Moreover, the learnable weight $\theta_{\mathrm{u}}$ modulates the slope of the logistic function, which allows learning a sharper binary decision boundary for optimal ID-OOD separation. This ablation demonstrates the superiority of the uncertainty loss employed by \texttt{\model}.
% \vspace{-0.5em}

\begin{table}[h]
    \centering
    \small
    \vspace{-1em}
    \begin{tabular}{c|ccc}
    \hline
        $\mathcal{L}_\text{uncertainty}$ & FPR95 $\downarrow$  & AUROC $\uparrow$  &mAP $\uparrow$   \\
        \hline
        & \multicolumn{2}{c}{\small COCO / nuImages as OOD}\\
        \cline{2-3}
        \model w/o $\theta_{\mathrm{u}}$ & 64.06 / 85.31 & 83.15 / 69.67 &30.1\\
      Hinge loss~\cite{liu2020energy} & 74.73 / 90.70   & 74.32 / 62.59  &30.2\\
      K+1 class   &84.34 / 93.63   &59.40 / 56.25  & 30.8\\
      \rowcolor{Gray}  \model (ours)& \textbf{52.18} / \textbf{77.57}& \textbf{85.67} / \textbf{75.67}&30.5 \\
         \hline
    \end{tabular}
    \vspace{-1em}
    \caption{\small Ablation study on the uncertainty regularization loss. }
    \label{tab:ablation_loss}
    % \vspace{-2em}
\end{table}

% we find that a learnable $\mathbf{w}$ for energy score is more desirable than a constant $\*w$, given the inherent class imbalance in object detection datasets. {Additional visualization of the learned weight coefficient $w_k$ is shown in Appendix~\ref{sec:app_visual_weight}, which aligns with the class frequency}. 

% Finally, 

% Comparison with hinge loss, additional K+1 class for ood class.

\subsection{Qualitative analysis}
%%%%%%%%% REFERENCES
\label{sec:visualization}
Here we further present   qualitative analysis on the instance-level OOD detection results. In Figure~\ref{fig:visual}, we visualize the predictions on several OOD images, using object detection models trained without distilled unknown objects (top) and with \texttt{\model} (bottom). The ID data is BDD100K. \texttt{\model} performs better in identifying OOD objects (in green) than a vanilla object detector and reduces false positives among detected objects. Moreover, the confidence score of the false-positive objects of \texttt{\model} is lower than that of the vanilla model (e.g., rocks in the 3rd column).
 \vspace{-0.5em}
% \begin{figure*}[t]
%     \centering
%     \includegraphics[width=1.0\textwidth]{figs/visual_crop.pdf}
%     \vspace{-1.5em}
%     \caption{\small Visualization of detected objects on the OOD images (from MS-COCO and NuImages) by a vanilla Faster-RCNN (\emph{top}) and \texttt{\model} (\emph{bottom}). The in-distribution is BDD-100k dataset. \textbf{Blue}: Objects detected and classified as one of the ID classes. \textbf{Green}: OOD objects detected by \texttt{\model}, which reduce false positives among detected objects.}
% % \vspace{-2em}
% \vspace{-1.5em}
% \label{fig:visual}
% \end{figure*}

\section{Related work}

\noindent\textbf{OOD detection for classification} can be broadly categorized into post hoc, generative-based and outlier exposure (OE)-based approaches~\cite{yang2021oodsurvey}. For {post hoc methods}, the softmax confidence score is a common baseline~\cite{hendrycks2016baseline}, which can be arbitrarily high for OOD inputs~\cite{hein2019relu}. Several improvements have been proposed, such as ODIN~\cite{liang2018enhancing, hsu2020generalized}, Mahalanobis~\cite{lee2018simple}, energy score~\cite{liu2020energy, wang2021canmulti}, Gram matrices
 score~\cite{DBLP:conf/icml/SastryO20}  and GradNorm score~\cite{huang2021importance}.\@
%Recent work explored self-supervised learning for OOD detection~\cite{mohseni2020self, tack2020csi}. 
%Different from~\cite{lee2018simple}, \texttt{VOS} performs dynamic estimation of class-conditional Gaussian during training, which shapes the uncertainty surface over time using a novel uncertainty regularization loss. 
{Outlier exposure} methods exploited regularization using natural images \cite{mohseni2020self,DBLP:journals/corr/abs-2106-03917,hendrycks2018deep,DBLP:conf/nips/DhamijaGB18,DBLP:conf/cvpr/LiV20,DBLP:journals/corr/abs-2108-11941}
%DBLP:journals/corr/abs-1906-03509
or images synthesized by GANs~\cite{lee2018training}.
%~\cite{DBLP:conf/nips/HendrycksMKS19,mohseni2020self, tack2020csi, sehwag2021ssd} and neural linear models~\cite{DBLP:conf/iclr/RiquelmeTS18}.
%and outlier selection~\cite{chen2020}. 
However, real outlier data is difficult to obtain, especially for object detection. In contrast, \texttt{\model} automatically distills unknowns from videos which allows greater flexibility.\@ 
%~\citet{DBLP:journals/corr/abs-2010-05119} also synthesized outliers in the feature space. It is different from \texttt{VOS} in the outlier generation method (sampling from class-conditional multivariate normal vs. normal), uncertainty score (generalized energy score vs. SVM outputs) and model architecture.
% Moreover, it stacked multiple autoencoders with two-level feature distillation, which is expensive for object detectors with high-resolution inputs.
{Generative models} directly estimate
the ID density~\cite{DBLP:conf/nips/SchirrmeisterZB20,DBLP:conf/nips/KingmaD18,van2016conditional},
%DBLP:conf/nips/KingmaD18, , van2016conditional
which makes them natural alternatives %tabak2013family
for OOD detection. 
However, they are in general less  competitive compared to discriminative-based methods and typically harder to optimize~\cite{DBLP:conf/iclr/HinzHW19,kirichenko2020normalizing, 
DBLP:conf/iclr/NalisnickMTGL19, DBLP:conf/nips/RenLFSPDDL19,DBLP:conf/iclr/SerraAGSNL20, xiao2020likelihood}. Very recently, Sun \emph{et al.}~\cite{sun2021react} showed that a simple activation rectification strategy termed ReAct can significantly improve test-time OOD detection.~Theoretical understandings on different post-hoc OOD detection methods are provided in~\cite{morteza2022provable}.~\cite{tack2020csi, sehwag2021ssd} applied self-supervised learning for OOD detection, which we compare in Section~\ref{sec:baselines}.
% \vspace{-0.4cm}

\noindent\textbf{OOD detection for object detection} is currently underexplored. Du \emph{et al.}~\cite{du2022vos} proposed to synthesize virtual outliers in the feature space for effective model regularization, and demonstrated promise on OOD detection for object detection.  In this paper, \texttt{\model} focuses on OOD detection with the help of videos and adopts an unknown-aware training loss. Moreover,~\cite{DBLP:journals/corr/abs-2103-02603} used the negative objects as the unknown samples, which is suboptimal as we show in Table~\ref{tab:synthesis_ablation}.
Harakeh \emph{et al.}~\cite{DBLP:journals/corr/abs-2101-05036} focused on uncertainty estimation for the localization branch, rather than OOD detection for classification problem. Several works~\cite{DBLP:conf/wacv/DhamijaGVB20,DBLP:conf/icra/MillerDMS19, DBLP:conf/icra/MillerNDS18,DBLP:conf/wacv/0003DSZMCCAS20,DBLP:journals/corr/abs-2108-03614} used approximate Bayesian methods, such as MC-Dropout~\cite{gal2016dropout} for OOD detection. They require multiple inference passes to generate the uncertainty score, which are computationally expensive on larger datasets and models. 
%
% with a new test-time detection score

%Our framework is the first to introduce an OOD-uncertainty-aware training for object detection, which shows superior performance compared to prior works.  %Both did not realize the importance of virtual outlier synthesis and are used as baselines for comparison with \texttt{VOS}.

\noindent\textbf{Open-world object detection} includes OOD generalization~\cite{DBLP:journals/corr/abs-2108-06753}, zero-shot object detection~\cite{DBLP:journals/corr/abs-2104-13921, DBLP:journals/ijcv/RahmanKP20}
%DBLP:conf/aaai/LiYZ0KZ19
and incremental object detection~\cite{DBLP:journals/corr/abs-2002-05347,DBLP:conf/cvpr/Perez-RuaZHX20}, etc. Generally they developed measures to mitigate catastrophic forgetting~\cite{DBLP:journals/corr/abs-2003-08798} or used auxiliary information~\cite{DBLP:journals/ijcv/RahmanKP20}, such as class attributes, to perform object detection on unseen data--both differing from our focus of OOD detection. Wang \emph{et al.}~\cite{DBLP:journals/corr/abs-2104-08381} adopted dissimilarity measurement in the cycle forward step, but their focus is OOD generalization (label space remains the same) rather than OOD detection. Additionally, it did not consider aggregating temporal information from multiple frames. 
%Our primary focus is OOD detection rather than open-set object detection. %While our primary focus is OOD detection, we believe \texttt{VOS} can benefit open-world detection tasks. 
 
 %measures of (spatial and semantic) uncertainty in object detectors to reject unknown classes~\citep{DBLP:conf/icra/MillerDMS19, DBLP:conf/icra/MillerNDS18}

\vspace{0.2cm}
 \noindent\textbf{Video anomaly detection (VAD)} aims to identify anomalous events on both the object level~\cite{DBLP:conf/cvpr/IonescuKG019,DBLP:conf/cvpr/DoshiY20b,DBLP:conf/mm/YuWCZXYK20} and frame level~\cite{DBLP:conf/cvpr/LiuLLG18,DBLP:conf/cvpr/MehranOS09,DBLP:conf/wacv/RavanbakhshNMSS18} by techniques such as skeleton trajectory modeling~\cite{DBLP:conf/cvpr/MoraisL0SMV19}, weakly supervised learning~\cite{DBLP:conf/eccv/ZaheerMAL20}, attention~\cite{DBLP:conf/cvpr/ParkNH20}, temporal pose graph~\cite{DBLP:conf/cvpr/MarkovitzSFZA20}, self-supervised learning~\cite{DBLP:conf/cvpr/GeorgescuBIKPS21} and autoencoders~\cite{DBLP:conf/eccv/ChangTXY20}. Compared with \texttt{\model}, the anomalies in VAD do not necessarily have different semantics from the ID training data. Moreover, none of the approaches considered synthesizing unknowns with the help of videos  or energy-based model regularization.
%  \vspace{-1em}
%  produce uncertainty estimates in a frame level rather than for instance-level predictions for object detection.

\section{Conclusion}
%  \vspace{-0.5em}
In this paper, we propose \texttt{\model}, an unknown-aware object detection framework for OOD detection. \texttt{\model} distills diverse unknown objects during training by exploiting the rich spatial-temporal information from videos. The distilled unknowns meaningfully improve the decision boundary between the ID and OOD data, resulting in state-of-the-art OOD detection performance while preserving the performance of the ID task. We hope our work will inspire future research towards unknown-aware deep learning in real-world settings.

\section{Acknowledgement}
Research is supported by Wisconsin Alumni Research Foundation (WARF), Facebook Research
Award, and funding from Google Research.

{\small
\bibliographystyle{ieee_fullname}
\bibliography{egbib}
}

\newpage
\onecolumn
\appendix

\begin{center}
    \Large{\textbf{Supplementary Material}}
\end{center}

\section{Experimental details}
We summarize the OOD detection evaluation task in Table~\ref{tab:task}. The OOD test dataset is selected from MS-COCO and nuImages dataset, which contains disjoint labels from the respective ID dataset. For the Youtube-VIS dataset, we use the dataset released in year 2021. Since there are no ground truth labels available for the validation images, we select the last 597 videos in the training set as the in-distribution evaluation dataset. The remaining 2,388 videos are used for training. The BDD100K and Youtube-VIS model are both trained for a total of 52,500 iterations. See {detailed ablations on the hyperparameters in Section 4.3 of the main paper}.

\begin{table}[!htb]
\centering
\begin{tabular}{@{}lrr@{}}
\toprule
 & \textbf{Task 1} &\textbf{ Task 2}  \\ \midrule
ID train dataset & BDD100K train & Youtube-VIS train \\
ID val dataset & BDD100K val & Youtube-VIS val  \\
OOD dataset & {COCO / nuImages} & {COCO  / nuImages}  \\
% Outlier dataset& \underline{OI val} & \underline{OI val} \\
$\#$ID train images & 273,406 & 67,861 \\
$\#$ID val images & 39,973 & 21,889\\
$\#$OOD images from COCO& 1,914 &28,922 \\
$\#$OOD images from nuImages& 2,100 & 2,100 \\ 
% $\#$OOD images & 969 & 1,914 \\ 
% $\#$Outlier images & 1,852 & 1,852  \\
% $\#$iterations (train)& 18,000& 90,000 \\
\bottomrule
\end{tabular}
\caption{OOD detection evaluation tasks. 
%The dataset with underline means a subset to maintain a disjoint class space.
}
\label{tab:task}
\end{table}

\section{In-distribution classes}
We provide a detailed description of the in-distribution classes for the two video datasets as follows.

BDD100K dataset contains 8 classes, which are  \emph{pedestrian, rider, car, truck, bus, train, motorcycle, bicycle}.
        
The Youtube-VIS dataset contains 40 classes, which are \emph{airplane, bear, bird, boat, car, cat, cow, deer, dog, duck, earless\_seal, elephant, fish, flying\_disc, fox, frog, giant\_panda, giraffe, horse, leopard, lizard, monkey, motorbike, mouse, parrot, person, rabbit, shark, skateboard, snake, snowboard, squirrel, surfboard, tennis\_racket, tiger, train, truck, turtle, whale, zebra}.

\section{Software and hardware}
\label{sec:hardware}
We run all experiments with Python 3.8.5 and PyTorch 1.7.0, using NVIDIA GeForce RTX 2080Ti GPUs.

% \section{Baselines}
\section{Baselines}
\label{sec:reproduce_baseline}
% In this section, firstly we disclose the implementation details for extending classic OOD detection baselines from classification to object detection. For all the baselines and our \texttt{VOS}, we do not sub-sample the detected objects to report metrics.

To evaluate the baselines, we follow the original methods in MSP~\cite{hendrycks2016baseline}, ODIN~\cite{liang2018enhancing}, Generalized ODIN~\cite{hsu2020generalized}, Mahalanobis distance~\cite{lee2018simple}, CSI~\cite{tack2020csi}, energy score~\cite{liu2020energy}  and gram matrices~\cite{DBLP:conf/icml/SastryO20} and apply them accordingly on the classification branch of the object detectors. For ODIN~\cite{liang2018enhancing}, the temperature is set to be $T=1000$ following the original work. For both ODIN and Mahalanobis distance~\cite{lee2018simple}, the noise magnitude is set to $0$ because the region-based object detector is not end-to-end differentiable given the existence of region cropping and ROIAlign.\@
%\@For Mahalanobis distance, we use 2000 examples randomly selected from the in-distribution training set and adversarial examples to train the logistic regression model. Same as in~\cite{lee2018simple}, the tuning set is generated by adding adversarial noise. The perturbation magnitude $\epsilon$ is chosen from
%\{0.0, 0.01, 0.005, 0.002, 0.0014, 0.001, 0.0005\}. The optimal parameters are chosen to minimize the FPR at
%TPR $95\%$, which is set to 0.0005.\@
%Note that for Generalized ODIN,\@
%we use inner-product (I) for the $h(\*x)$ head which yields the best performance among three variants.\@
%we do not apply input processing on the test images since we do not observe performance benefit. 
For GAN~\cite{lee2018training}, we follow the original paper and use a GAN to generate OOD images. The prediction of the OOD images/objects is regularized to be close to a uniform distribution, through a KL divergence loss with a weight of 0.05.
%Since there are no ground truth bounding boxes for the generated fake images, we re-use the proposals and the ground-truth bounding boxes on the fake images. 
We set the shape of the generated images to be 100$\times$100 and resize them to have the same shape as the real images. We optimize the generator and discriminator using the Adam optimizer~\cite{DBLP:journals/corr/KingmaB14}, with a learning rate of 0.001. 
For CSI~\cite{tack2020csi}, we use the rotations (0$^\circ$, 90$^\circ$, 180$^\circ$, 270$^\circ$) as the self-supervision task. We set the temperature in the contrastive loss to 0.5. We use the features right before the classification branch (with the dimension to be 1024) to perform contrastive learning. The weights of the losses that are used for classifying shifted instances and instance discrimination are both set to 0.1 to prevent training collapse. For Generalized ODIN~\cite{hsu2020generalized}, we replace and train the classification head of the object detector by the most effective Deconf-C head shown in the original paper.

\begin{table*}[!htb]
\centering
\scalebox{0.8}{
\begin{tabular}{lllll}
    \toprule
    {In-distribution $\mathcal{D}$}  & 
     
   {\textbf{Method }}  &\textbf{FPR95} $\downarrow$  & \textbf{AUROC} $\uparrow$ & \textbf{ mAP (ID)}$\uparrow$  \\
   \hline
    && \multicolumn{2}{c}{OOD: MS-COCO / nuImages}\\
     \cline{3-4}
    \multirow{9}{0.2\linewidth}{{\textbf{ BDD100K} }} 
    & MSP \cite{hendrycks2016baseline}
   &80.09 / 93.05 & 74.19 / 63.14& 32.0 \\
     & ODIN \cite{liang2018enhancing}
    &64.74 / 82.08 & 77.65 / 67.09  &32.0 \\
     & Mahalanobis 
    \cite{lee2018simple}
    &  54.02 / 79.85   & 82.38 / 75.48& 32.0\\ 
        & Gram matrices~\cite{DBLP:conf/icml/SastryO20}& 63.96 / 63.61    & 67.56 / 67.47 	&32.0 \\
      & Energy score~\cite{liu2020energy}
    & 64.79 / 81.62 &  78.78 / 69.43  & 32.0\\ 
    % & Entropy score~\cite{DBLP:journals/corr/abs-2101-05036} & 68.48&	71.65 & 49.2 \\\
     & Generalized ODIN~\cite{hsu2020generalized} & 60.76 / 82.00 & 80.14 /  70.74 &32.5
     \\
  & CSI~\cite{tack2020csi}&52.98 / 80.00   & 83.57 / 74.91& 31.8\\
    & GAN-synthesis~\cite{lee2018training}&58.35 / 83.65  & 81.43 /  70.39 & 31.5\\
    %  & SS-OOD~\cite{DBLP:conf/nips/HendrycksMKS19} &59.68 & 83.08  & 48.4\\
    %  & \rowcolor{Gray} \textbf{\model} w/ Pair-frame  (ours) 
    % & 
    
    % 61.41^{\pm1.0} / 82.24^{\pm1.3}  &83.29^{\pm0.3}	/ 74.34^{\pm0.4}&  30.6^{\pm0.1}\\
     & \cellcolor{Gray} \textbf{\model}  (ours)  &\cellcolor{Gray}\textbf{52.51 / 79.75 }  & \cellcolor{Gray}\textbf{84.03 / 76.55} &  \cellcolor{Gray}32.3 
     \\
%       & \rowcolor{Gray} \textbf{\model}  (ours)  &\textbf{}^{\pm} / \textbf{}^{\pm} & \textbf{}^{\pm} / \textbf{}^{\pm}&
% ^{\pm}  \\
     \midrule
 \multirow{10}{0.2\linewidth}{\textbf{Youtube-VIS} } 
     & MSP \cite{hendrycks2016baseline}
    & 89.86 / 97.42& 67.04 / 54.02& 26.7 \\
     & ODIN \cite{liang2018enhancing}
    & 89.28 / 96.30 & 67.54 / 60.82 & 26.7 \\
     & Mahalanobis 
    \cite{lee2018simple}
   	&90.00 / 94.44 & 70.47 / 54.83  & 26.7\\ 
   	    & Gram matrices~\cite{DBLP:conf/icml/SastryO20}& 87.64 / 91.25 & 69.76 / 61.43	&26.7  \\
      & Energy score~\cite{liu2020energy}
    &88.54 / 90.21& 67.83 / 58.02& 26.7\\ 
    % & Entropy score~\cite{DBLP:journals/corr/abs-2101-05036} & 54.28 & 79.42 &  31.8 \\
      &Generalized ODIN~\cite{hsu2020generalized} &85.15 / 98.00  & 71.57 / 64.23   &27.3
     \\
     &  CSI~\cite{tack2020csi}& 82.43 / 88.61  & 71.81 / 54.00 &24.2\\
    & GAN-synthesis~\cite{lee2018training}&	85.75 / 93.75  & 72.95 / 56.94 &25.5\\
    % & SS-OOD~\cite{DBLP:conf/nips/HendrycksMKS19} &49.39 &84.54 &30.6 \\
    %  & \rowcolor{Gray} \textbf{\model} w/ pair-frame   (ours) 
    % & 	81.81^{\pm0.7} / 82.13^{\pm0.8} & 73.62	^{\pm0.5} / 70.21^{\pm0.8}	&  24.8^{\pm0.0}  \\
     & \cellcolor{Gray} \textbf{\model}  (ours) 
    & \cellcolor{Gray}\textbf{81.14} / \textbf{80.77}&\cellcolor{Gray}\textbf{74.82} / \textbf{69.52}&\cellcolor{Gray}27.2  \\
%           & \rowcolor{Gray} \textbf{\model}  (ours)  &\textbf{}^{\pm} / \textbf{}^{\pm} & \textbf{}^{\pm} / \textbf{}^{\pm}&
% ^{\pm}  \\
    %  & \rowcolor{Gray} \textbf{VOS}-RegX4.0 (ours)  &\textbf{38.79}^{\pm1.3} / \textbf{28.67}^{\pm1.7}  &\textbf{88.21}^{\pm0.9} / \textbf{91.05}^{\pm0.4}& \textbf{32.5}^{\pm0.0}
        \bottomrule
\end{tabular}}
        \vspace{-0.2cm}
        \caption[]{\small Comparison with competitive out-of-distribution detection methods. All baseline methods are based on a model trained on {ID data only} using RegNetX-4.0GF as the backbone.  $\uparrow$ indicates larger values are better, and $\downarrow$ indicates smaller values are better. All values are percentages. \textbf{Bold} numbers are superior results.  }
        \label{tab:app_baseline}
        \vspace{-0.8em}
\end{table*}

\section{Ablation study on a different backbone architecture}
\label{sec:backbone_app}
In this section, we evaluate the proposed \model using a different backbone architecture of the Faster-RCNN, which is RegNetX-4.0GF~\cite{DBLP:conf/cvpr/RadosavovicKGHD20}. Similarly, we compare with the same set of OOD detection baselines as stated in the main paper. The results are shown in Table~\ref{tab:app_baseline}. 

From Table~\ref{tab:app_baseline}, we demonstrate that \model
is effective on alternative neural network architectures. In particular, using RegNet~\cite{DBLP:conf/cvpr/RadosavovicKGHD20} as backbone yields better OOD detection performance compared with the baselines. 
Moreover, we show that \texttt{\model} achieves stronger OOD detection performance while preserving or even slightly increasing the object detection accuracy on ID data (measured by mAP). This is in contrast with CSI, which displays significant degradation, with mAP decreasing by 3\% on Youtube-VIS.
%Meanwhile, the  ID accuracy is improved compared with the backbone architecture of ResNet-50, i.e., achieving a mAP of 32.3\% vs. 30.5\% on the BDD100K dataset.

\section{Additional visualization examples}
\label{sec:app_visual}
\begin{figure}[h]
    \centering
    \includegraphics[width=1.0\textwidth]{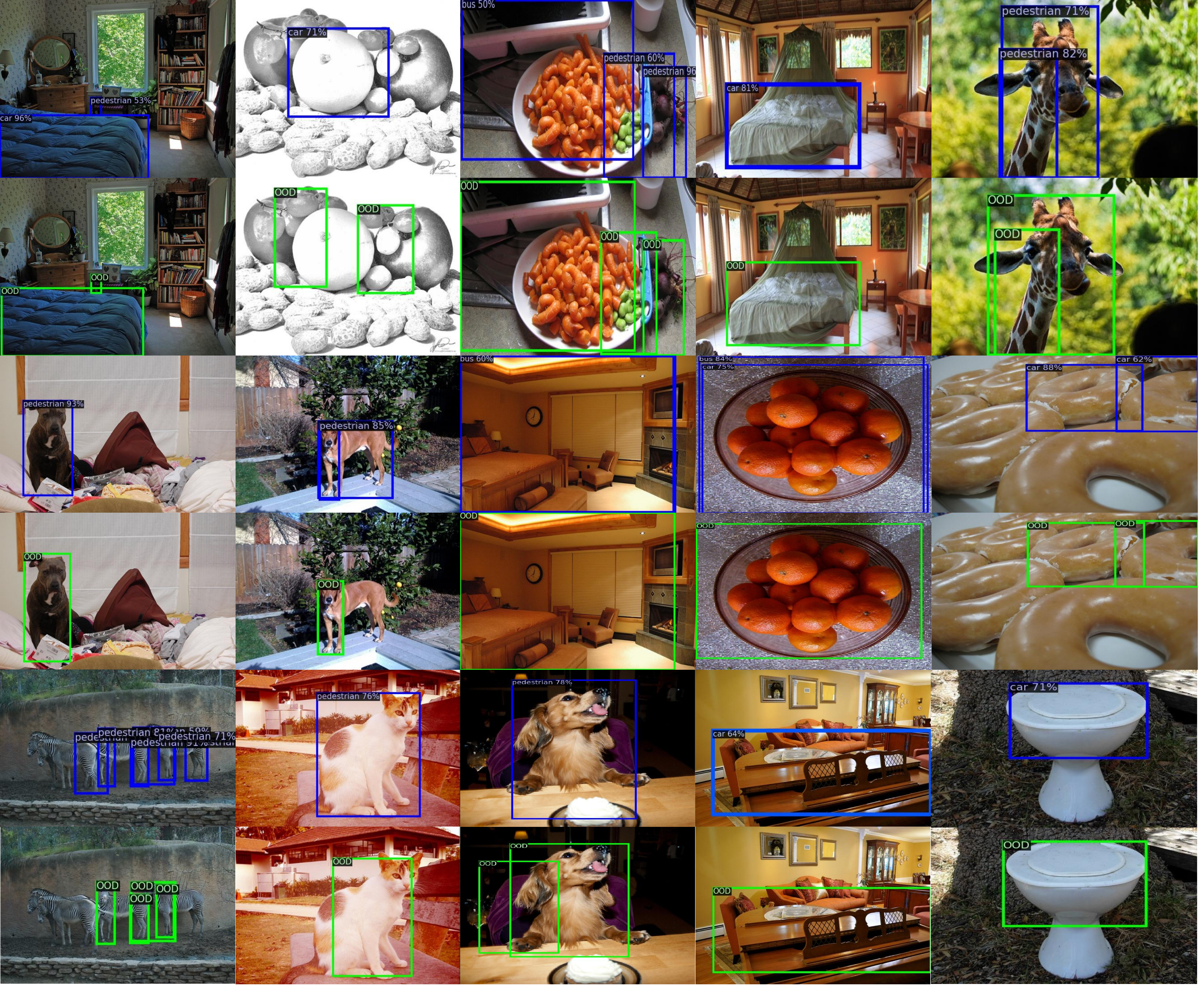}
    % \vspace{-1.5em}
    \caption{\small Additional visualization of detected objects on the OOD images (from MS-COCO) by a vanilla Faster-RCNN (\emph{top}) and \model (\emph{bottom}). The in-distribution is BDD100K dataset.  \textbf{Blue}: Objects detected and classified as one of the ID classes. \textbf{Green}: OOD objects detected by \model, which reduce false positives among detected objects.}
% \vspace{-2em}
% \vspace{-2em}
\label{fig:vi1}
\end{figure}

\begin{figure}[h]
    \centering
    \includegraphics[width=1.0\textwidth]{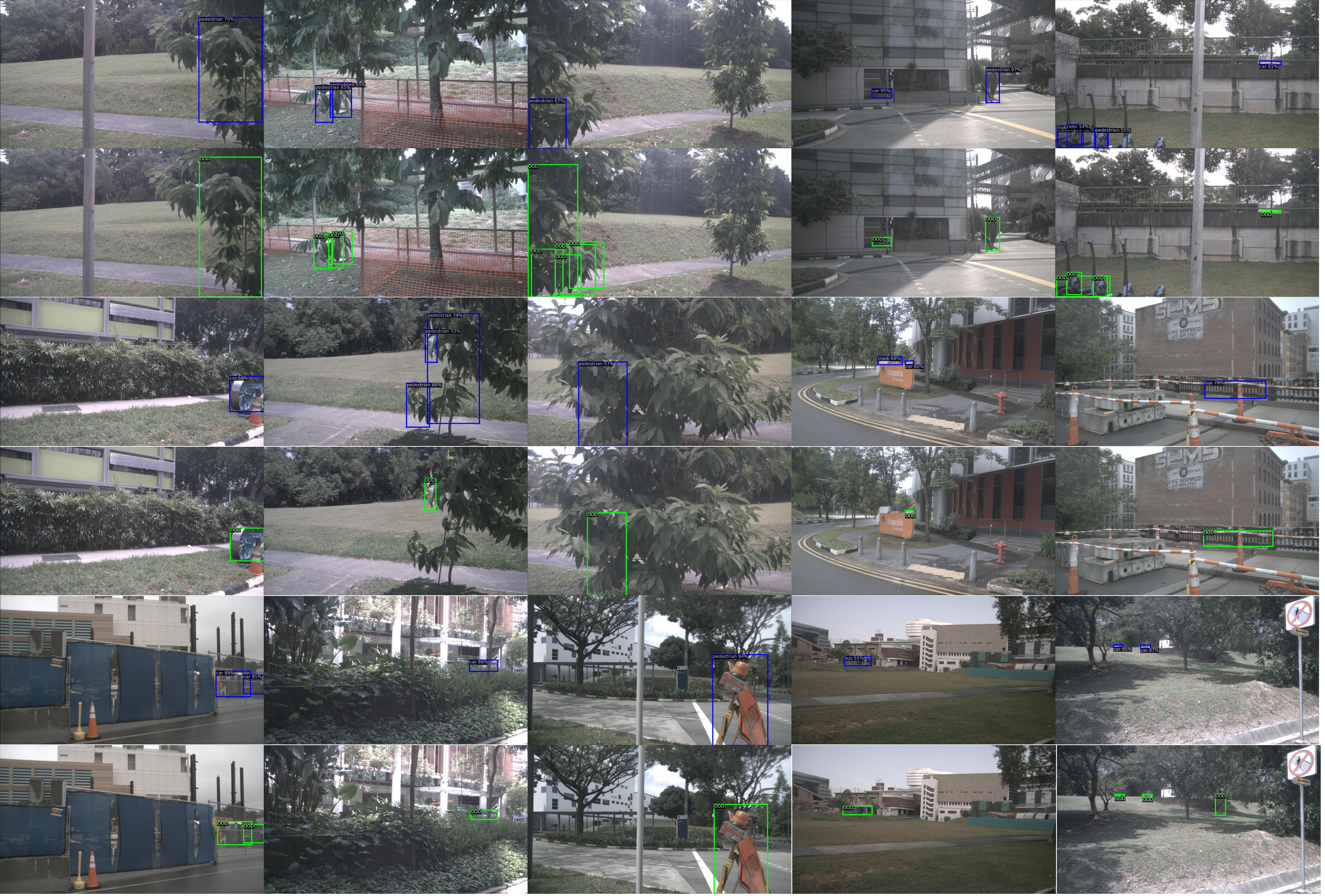}
    % \vspace{-1.5em}
    \caption{\small Additional visualization of detected objects on the OOD images (from nuImages) by a vanilla Faster-RCNN (\emph{top}) and \model (\emph{bottom}). The in-distribution is BDD100K dataset.  \textbf{Blue}: Objects detected and classified as one of the ID classes. \textbf{Green}: OOD objects detected by \model, which reduce false positives among detected objects.}
% \vspace{-2em}
% \vspace{-2em}
\label{fig:vi2}
\end{figure}

\begin{figure}[h]
    \centering
    \includegraphics[width=1.0\textwidth]{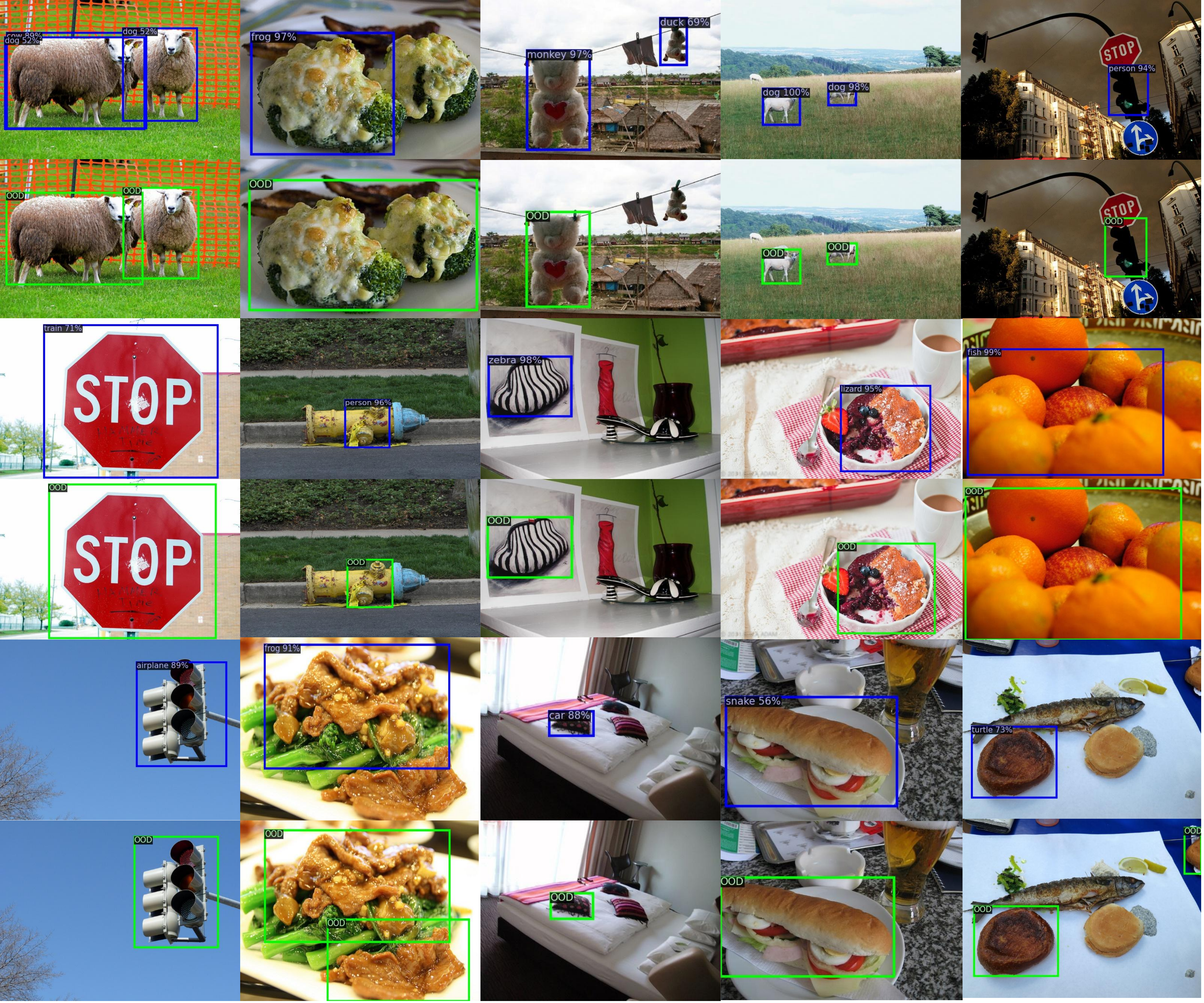}
    % \vspace{-1.5em}
    \caption{\small Additional visualization of detected objects on the OOD images (from MS-COCO) by a vanilla Faster-RCNN (\emph{top}) and \model (\emph{bottom}). The in-distribution is Youtube-VIS dataset.  \textbf{Blue}: Objects detected and classified as one of the ID classes. \textbf{Green}: OOD objects detected by \model, which reduce false positives among detected objects.}
% \vspace{-2em}
% \vspace{-2em}
\label{fig:vi3}
\end{figure}

\begin{figure}[h]
    \centering
    \includegraphics[width=1.0\textwidth]{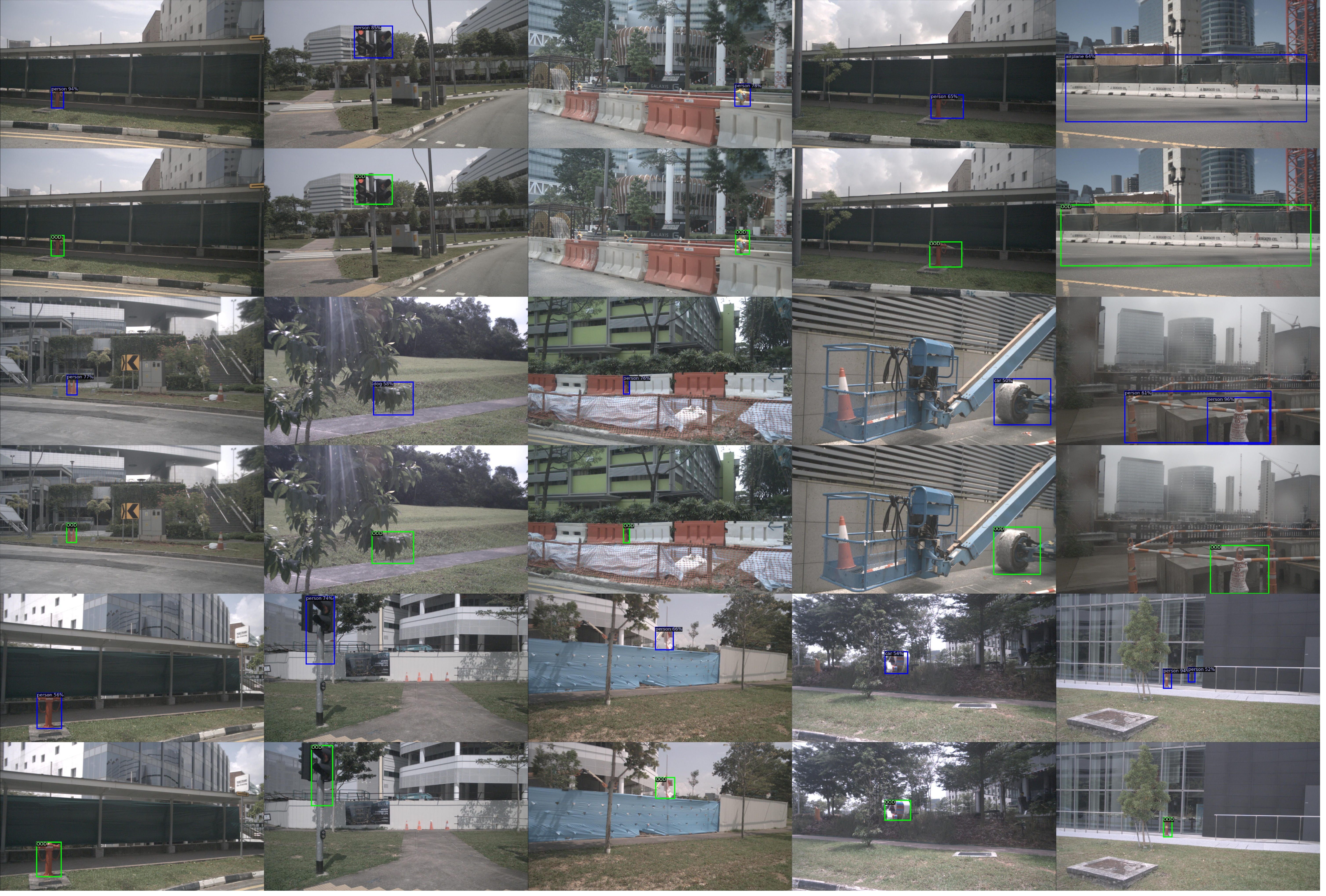}
    % \vspace{-1.5em}
    \caption{\small Additional visualization of detected objects on the OOD images (from nuImages) by a vanilla Faster-RCNN (\emph{top}) and \model (\emph{bottom}). The in-distribution is Youtube-VIS dataset.  \textbf{Blue}: Objects detected and classified as one of the ID classes. \textbf{Green}: OOD objects detected by \model, which reduce false positives among detected objects.}
% \vspace{-2em}
% \vspace{-2em}
\label{fig:vi4}
\end{figure}

We provide additional visualization of the detected objects on different OOD datasets with models
trained on different in-distribution datasets. The results are shown in Figures~\ref{fig:vi1}-\ref{fig:vi4}.

% \section{Limitations}
% Our \model performs unknown distillation in the video datasets, which is less applicable to object detection datasets with still images. However, this limitation can be mitigated by setting the frame sampling range to infinity where the selected key frames and reference frames do not have a temporal coherence constraint.

% {\small
% \bibliographystyle{ieee_fullname}
% \bibliography{egbib}
% }

% \end{document}

\end{document}